\documentclass[12pt]{iopart}
\usepackage[utf8]{inputenc} 
\usepackage[T1]{fontenc}    
\usepackage{hyperref}       
\usepackage{url}            
\usepackage{booktabs}       
\usepackage{amsfonts}       
\usepackage{nicefrac}       

\usepackage{amsmath}
\makeatletter
\@namedef{ver@amsmath.sty}{}
\makeatother
\usepackage{amstext}
\usepackage{chemformula}

\usepackage{savesym}
\usepackage{microtype}      
\usepackage{algorithm}
\savesymbol{equation*}
\usepackage[linesnumbered,ruled,vlined,algo2e]{algorithm2e}
\usepackage{algpseudocode}
\usepackage{amssymb}
\usepackage{graphicx}
\usepackage{xcolor}

\newcommand{\bx}{\mathbf{x}}
\newcommand{\bu}{\mathbf{u}}
\newcommand{\bv}{\mathbf{v}}
\newcommand{\bg}{\mathbf{g}}
\newcommand{\bW}{\mathbf{W}}
\newcommand{\bw}{\mathbf{w}}
\newcommand{\bT}{\mathbf{T}}
\newcommand{\bn}{\mathbf{n}}
\newcommand{\xCM}{x^\text{\tiny CM}}
\newcommand{\bxCM}{\mathbf{x}^\text{\tiny CM}}
\newcommand{\rCM}{r}

\begin{document}

\title[Averaged Gradient Descent]{How to iron out rough landscapes and get optimal performances: 
Averaged Gradient Descent \\
and its application to tensor PCA}

\author{Giulio Biroli}

\address{Laboratoire de Physique de l'Ecole Normale Sup\'erieure ENS, Universit\'e PSL, CNRS, Sorbonne Universit\'e, Universit\'e Paris-Diderot, Sorbonne Paris Cit\'e, Paris, France}
\ead{giulio.biroli@ens.fr}

\author{Chiara Cammarota}

\address{Department of Mathematics,
  King's College London,
  Strand, London WC2R 2LS}
\ead{chiara.cammarota@kcl.ac.uk}

\author{Federico Ricci-Tersenghi}

\address{Dipartimento di Fisica, Sapienza Universit\`a di Roma,
  INFN -- Sezione di Roma1, and CNR-Nanotec, unit\`a di Roma,
  P.le A. Moro 5, Roma 00185 Italy}
\ead{federico.ricci@uniroma1.it}

\vspace{10pt}

\begin{abstract}
In many high-dimensional estimation problems the main task consists in minimizing a cost function, which is often strongly non-convex when scanned in the space of parameters to be estimated.
  A standard solution to flatten the corresponding rough landscape consists in summing the losses associated to different data points and obtain a smoother empirical risk. Here we propose a complementary method that works for a single data point. 
  The main idea is that a large amount of the roughness is uncorrelated in different parts of the landscape. 
  One can then substantially reduce the noise by evaluating an empirical average of the gradient obtained as a sum over many random independent positions in the space of parameters to be optimized. 
  We present an algorithm, called Averaged Gradient Descent, based on this idea and we apply it to tensor PCA, which is a very hard estimation problem. 
  We show that Averaged Gradient Descent over-performs physical algorithms such as gradient descent and approximate message passing and matches the best algorithmic thresholds known so far, obtained by tensor unfolding and methods based on sum-of-squares.  
\end{abstract}

%
%
%
%
%

\section{Introduction}
One recurrent central task in many modern machine learning problems is the minimization of a non-convex high-dimensional function. 
Gradient descent is a versatile workhorse method that is widely used in these contexts, in particular in high-dimensional estimation to optimize the  likelihood function. 
However the performance of gradient descent can be substantially undermined in cases where the function to be optimized---or informally the landscape---is rough. One way out is to increase the signal to noise ratio by summing the losses associated to different data points and obtain a smoother empirical risk.\\ In this work we propose an alternative method which works for a single data point. Our main idea is that a large amount of the roughness is uncorrelated in different parts of the landscape. By evaluating an empirical average of the gradient obtained as a sum over many random independent positions in the space of parameters to be optimized, 
one can then substantially reduce the noise, thus effectively ironing out the landscape and letting the signal contribution emerge.
\\ 
We propose an algorithm, called  Averaged Gradient Descent (AGD), based on this idea. We test it on tensor-PCA \cite{MonRic14}, a very hard high-dimensional estimation problem in which
one observes a $k$-fold $N\times N\times \cdots \times N$ tensor
\begin{equation}
{\bT} = {\bW} + \frac{\lambda}{N^{\frac{k-1}{2}}} {\bv}^{\otimes k}\;,
\end{equation}
where ${\bW}$ is a symmetric noise tensor with independent normally distributed elements and $\lambda$ represents the signal to noise ratio (SNR).
The aim is to recover the signal ${\bv}\in \mathbb R^N$ with $||\bv||_2=\sqrt{N}$. Without loss of generality one can take ${\bv}$ pointing in a random direction on the surface $S^{N-1}(0,\sqrt{N})$ of an hyper-sphere of radius $\sqrt{N}$ centred in the origin.
The Maximum Likelihood estimate of ${\bv}$ is the vector ${\bx^*}$ of norm $\sqrt{N}$ that minimizes 
$
\sum_{i_1\le\dots\le i_k} \left(T_{i_1,\dots,i_k}- x_{i_1}\dots x_{i_k}\right)^2
$, which leads to 
\begin{equation}
{\bx^*}={\arg\max}_{\bx \in S^{N-1}} \sum_{i_1\le\dots\le i_k} T_{i_1,\dots,i_k}x_{i_1}\dots x_{i_k} \ .
\end{equation}
From a statistical mechanics perspective the previous equation on ${\bx^*}$ can be also seen as the minimization equation  
of the following energy function:
\begin{align}
    H(\mathbf{x})&=-\frac{1}{N^{\frac{(k-1)}{2}}}\sum_{i_1\le\dots\le i_k}T_{i_1,\dots,i_k} x_{i_1}\dots x_{i_k}\\
    &=-\frac{1}{N^{\frac{(k-1)}{2}}}\sum_{i_1\le\dots\le i_k}W_{i_1,\dots,i_k} x_{i_1}\dots x_{i_k} - N \frac{\lambda}{k!} m^k
\end{align}
where $m=(\bv,\bx)/N=\sum_{i=1}^N v_i x_i/N$ is the overlap with the signal or the magnetization of configuration ${\bx}$ in statistical physics language.\\
It is known that when 
$\lambda>\lambda_\text{IT}(k)$ with $\lambda_\text{IT}(k)$ of order one in the large $N$ limit (e.g. $\lambda_\text{IT}(3)\simeq 2.955$) it is information theoretically possible to recover the signal \cite{MonRic14,lesieur2017statistical}. 
However, a much larger SNR, $\lambda\gg N^{\frac{k-2}{4}}$, has to be reached in order to find algorithms, such as tensor unfolding and the ones based on sum of squares \cite{MonRic14,HoShSt15}, able to recover the signal in polynomial time. Gradient descent, and other physical algorithms as Approximate Message Passing (AMP) and Langevin dynamics, 
are sub-optimal and succeed only for $\lambda\gg N^{\frac{k-2}{2}}$ \cite{MonRic14,BaGhJa18}.
The inefficiency of physical algorithms is conjectured to be related to the roughness of the energy landscape, which is characterized by an exponential number of minima in the band $m< m_{\rm tr}$, with $m_{\rm tr}$ shrinking to zero as an inverse power of $N$ for any $\lambda$ growing sub-exponentially with $N$ \cite{BeMeMN17,RoBeBC19} (see also SM). Tensor-PCA therefore provides a very good framework to test whether our method for ironing out the landscape using multiple uncorrelated copies is efficient and able to match the performance of the best algorithms not based on the landscape (e.g.\ spectral methods). We show that this is indeed the case. An additional outcome of our analysis is the demonstration by systematic numerical studies that the {\it algorithmic threshold} of AGD is associated to a {\it threshold phenomenon} (a phase transition in physics jargon) that we fully characterize. \\
Our approach is grounded on the research axis which aims to connect the behavior of dynamics and algorithms to landscape properties, and to exploit the knowledge of the latter to improve the performance of the former. In fact, as we shall show, AGD is an algorithm fully rooted on physical intuition and aimed at optimally exploiting the information gained from the analysis of the landscape. 
In matching the best algorithmic performances achieved so far for tensor-PCA, AGD re-establishes the competitiveness of landscape-based  algorithms originating from statistical physics and clarifies to what extent algorithmic transitions are determined by {\it landscape properties}. 
From a more general point of view, AGD inherits the versatility of gradient descent, and hence stands as a new efficient algorithm suitable to a very wide spectrum of applications.

\section{Related Works}

Different procedures have been devised to regularize a rough landscape and improve optimization performance. 
One approach is based on the convolution of a rough energy function with a smoothing kernel \cite{wu1996effective}. 
Another procedure is based on the introduction of different copies of the system which are coupled together \cite{BBCILSZ16}. In both cases, the idea is to reduce the roughness by smoothing the landscape {\it locally} on sets of points with high overlap. Our method, instead, 
aims at reducing the roughness by a much more {\it global} average over uncorrelated copies, which have typically zero overlap.\footnote{One may argue that Hamming distances between copies are extensive both in local and global methods, however copies in AGD are as spread as possible, while local methods must keep copies close enough.}\\
Among the many algorithms devised for tensor PCA, the one based on homotopy \cite{AnDeGM16} is the closest one to AGD, although it was introduced from a very different perspective. From a general point of view, the main difference is that our method can be straightforwardly applied and extended to generic high-dimensional inference problems. We will compare in more detail later in the main text and in SM8 the two methods and their performances in the context of tensor PCA, showing the superiority of AGD in the large $N$ limit.\\
Finally, we point out that the optimality gap between algorithms not based on the landscape and statistical physics methods was very recently bridged by an extension of approximate message passing based on the Kikuchi approximation \cite{wein2019kikuchi}. 
Our results show that the gap can be also closed by using an extension of gradient descent. 
In this way a full redemption \cite{wein2019kikuchi} of the landscape dominated statistical physics approach against sophisticated algorithms not based on the landscape is reached.

\section{Averaged Gradient Descent}
The approach we propose here aims at being completely general. It 
takes advantage of physical intuition for the construction of a simple gradient-descent-based algorithm able to navigate through rough landscapes, hence, reaching very good algorithmic performances.
Averaged Gradient Descent uses the simple idea that sampling several independent locations, called {\it real replicas} of the system, helps decreasing the roughness of the landscape which originates from uninformative corrupting noise. In fact, the average over the replicas leads to a relative amplification of the informative contribution produced by the signal with respect to the noise.  
Note that AGD can be generalized and potentially applied to the broad range of problems in high-dimensional inference ({\it i.e.} 
other tensor problems \cite{HilLim13}, 
compressed sensing \cite{Donoho06}, 
community detection \cite{DeKrMZ11,DeKrMZ11bis,AnDeGM16}, 
learning graphical models \cite{ChaLia14} just to mention a few examples) 
that, in certain regimes of the parameters, are characterized by a hard phase where uninformative spurious minima trap local dynamics and hamper the reconstruction of the signal.
As anticipated in the introduction, in what follows we enter in the details of the application of this new algorithm to tensor-PCA, which is a notoriously hard problem in this sense, 
and we comment on the possibility of its  generalization. 

\begin{algorithm}[H] 
\caption{Averaged Gradient Descent, AGD}
\DontPrintSemicolon 
\KwIn{Landscape $H({\bx})$, number of replicas $R$, learning rate $\eta$, stopping criterion $\varepsilon$}
\KwOut{Estimate of the location of the landscape minimum ${\bx^*}$}
 $t\gets 0;\quad \bxCM(0)\gets0; \quad \rCM(0) \gets 0$
 \tcp*{\small Initialize the center of mass}
\label{alg:AGD}
 \Repeat {$||\bxCM(t)-\bxCM(t-1)||_2<\varepsilon$}{
  \For(\tcp*[f]{\small Given center of mass, sample $R$ points on the sphere}){$\alpha=1,\ldots,R$}{
  $\bx_\alpha(t) \gets \bxCM(t) + \sqrt{1-\rCM^2(t)}\,\bu_\alpha(t)$
  with $\bu_\alpha(t)$ drawn uniformly at random among vectors such that $||\bu_\alpha(t)||_2^2=N$ and $(\bu_\alpha(t),\bxCM(t))=0$
  
  $\bg_\alpha(t)\gets\nabla H|_{\bx_\alpha(t)}$ \tcp*{\small Evaluate the gradient on each of the $R$ points ${\bx_\alpha}(t)$}
  }
  $\bxCM(t+1) \gets \bxCM(t) - \eta \sum_\alpha \bg_\alpha(t)/R$
  
  \tcp*{\small Use the average gradient to update the position of the centre of mass}
  
  $t \gets t+1$ 

  $\rCM(t) \gets ||\bxCM(t)||_2 / \sqrt{N}$
  
  \If(\tcp*[f]{\small Keep the centre of mass inside or on the sphere}){$\rCM(t)>1$}{
   $\bxCM(t) \gets \bxCM(t)/\rCM(t)$
   
   $\rCM(t)\gets 1$
   \tcp*{\small When $\rCM(t)=1$ the algorithm reduces to standard GD}
  }
 }(\tcp*[f]{\small stopping condition as in standard GD})
 \Return{$\bxCM(t)$}
\end{algorithm}

\noindent The heuristics behind this algorithm is very simple and can be discussed in full generality. For non zero signal to noise ratio $\lambda$, whenever the local information on the gradient ${\bg}_{\alpha}$ contains a tiny component $\lambda {\bg_{\mathbf{s},\alpha}}$ systematically pointing in the direction of the signal, this algorithm aims at getting it amplified with respect to the complementary component of the gradient that is originated by uninformative corrupting noise, ${\bg_{\bn,\alpha}}={\bg}_{\alpha}-\lambda {\bg_{\mathbf{s},\alpha}}$.\footnote{
Note that the possibility to linearly decompose the gradient into these two terms can be considered general in the limit of small $\lambda {\bg_{\mathbf{s},\alpha}}$, as in this limit each gradient can be expanded around the zero-signal limit and the expansion truncated to the first order. The same decomposition is straightforward in the case of tensor PCA.}
For a given sample and considering different configurations ${\bx}$ drawn at random on the sphere,   
${\bg_{\bn,\alpha}}({\bx})$ is expected to have a strong fluctuating part and a small average ${\bg_{\bn,\alpha}^{\rm av}}$.  
The Central Limit Theorem implies that averaging over $R$ independent replicas of the system leads to a suppression by a factor $1/\sqrt{R}$ of the fluctuating part. By these simple arguments we conclude that the averaged algorithm will end up operating under a much higher effective signal to noise ratio. 
However, above a certain number of replicas, $R_{\rm opt}$, the fluctuating part becomes subleading with respect to ${\bg_{\bn,\alpha}^{\rm av}}$ and one cannot iron out more the landscape. Thus, $R_{\rm opt}$ sets the optimal number of replicas that have to be used in practice. This number is evaluated for tensor PCA in the Supplementary Material (SM3).\\
The explanation above holds, and the proposed algorithm gives a net advantage in the retrieval of the signal, when the problem has only one optimal solution.
This situation corresponds for instance to the case of tensor PCA with $k$ odd. When two degenerate solutions are present, {\it e.g.} for tensor PCA with $k$ even or any other inference problem where the global sign of the solution does not really matter, the multiple sampling of the landscape at $t=0$ through independent different copies of the system will not be of any help. The reason is that the local gradient sampled through $R$ different replicas will be randomly pointing towards any of the two solutions and their average will be suppressed by a factor $1/\sqrt{R}$, {\it i.e.} exactly at the same pace as the uninformative component originated by the noise. 
In this case we suggest to replace the averaged gradient in Algorithm \ref{alg:AGD} by the eigenvector ${\bw}_{\rm min}(t)$, with norm $\sqrt{N}$, corresponding to the minimum eigenvalue of the averaged Hessian $\sum_{\alpha}\mathcal{H}_{ij}|_{{\bx}_{\alpha}(t)}/R$. This procedure can be seen, for $t=0$, as a new general way to obtain spectral methods for high-dimensional inference problems \cite{lu2017phase}.
An additional care is needed here to keep consistency, step by step, of the sense of the update vector. This issue is solved by asking that the scalar product is $({\bw}_{\rm min}(t),{\bw}_{\rm min}(t+1))>0 $. After a few steps $t^*$ the symmetry between the two solutions is broken therefore it is advisable to continue with the original algorithm based on gradients, which is less computationally expensive. \\
In the general case a good strategy is to compute for the first steps both the average gradient and the average Hessian with its lowest eigenvector. Among the two averaged vectors the one to be used is the one leading to larger decrease in the energy function. After few steps the average gradient should become larger and should point towards the signals (even in the symmetric case of even $k$), one can then continue with Algorithm \ref{alg:AGD}. \\
In the next sections we are going to focus more specifically on the performances of this algorithm on tensor PCA. 
Interestingly in this case
not only the analysis at finite $R$ can be performed but also the study at infinite $R$, 
which turns out to be even computationally convenient.
Therefore in what follows we are going to focus on the large $R$ limit of AGD, where empirical averages are substituted by expected values on the uniform measure over the space of variables. Such algorithm involving infinite real replicas is hereafter called iAGD (infinite-$R$ AGD). Its performance will be discussed in the section Numerical Results.
Note that to develop an analytic understanding of these results we resort to a further simplification of the algorithm as discussed in the next section.
Finally the results for finite $R$ will be quoted and explained in the Supplementary Material (SM7).  

\section{Theoretical analysis: From landscape properties to the performance of the simplest optimal algorithms}
So far, we have introduced AGD and its $R\rightarrow \infty$ version called iAGD. Both algorithms are very challenging to be fully analyzed. For this reason, in this section we introduce a simplified version, SiAGD, and present its full theoretical analysis which provides several insights on the behavior of AGD and iAGD. In the next sections and in the SM, we then confirm these results and fully analyze these two algorithms by numerical experiments.  \\
The key idea to simplify AGD and iAGD applied to tensor PCA is that both algorithms are characterized by two regimes: a first one where the norm of the centre of mass increases from zero to $\sqrt{N}$, and a second one which corresponds to simple gradient descent (when the center of mass reaches the surface of the sphere all replicas fall on the centre of mass). 
The simplified version that we analyze here, SiAGD, consists in modifying the first regime by 
moving straight in the direction of the $t=0$ (averaged) gradient until the centre of mass hits the hyper-sphere. 
We shall show that SiAGD has an algorithmic threshold  
for the recovery of the signal, which is optimal compared to the ones of all the other algorithms known so far. Its numerical analysis and a comparison with iAGD and AGD is presented later.  
We will consider separately the odd and even $k$ cases since the simplified algorithm
is different, actually even simpler in the even case. 
Moreover, focusing on SiAGD as a simpler version of iAGD, we will work directly with averaged quantities. However it should be kept in mind that they can be estimated accurately using empirical averages over a large enough number of real replicas as discussed at the end of this section and in the SM (SM3 and SM7). Finally, we will always consider that the rate $\eta$ is small enough so that the discrete updates in the algorithm can be considered a good approximation of a continuous time algorithm. 

\subsection{Case I: k odd}
The value of the averaged gradient at $t=0$ for iAGD
reads: 
$$
g_i=-\frac{1}{N^{\frac{(k-1)}{2}}}\sum_{i_2\le\dots\le i_k}W_{i,i_2\dots,i_k} \mathbb {E}[x_{i_2}\dots x_{i_k}] -  \frac{\lambda}{(k-1)!N^{k-1}} v_i \sum_{i_2,\cdots,i_k}
v_{i_2}\cdots v_{i_k}\mathbb {E}[x_{i_2}\dots x_{i_k}] \ .
$$
The expectation $\mathbb {E}[\cdot]$ is over the uniform measure on the sphere of radius $\sqrt N$.
SiAGD consists in doing GD by using this initial averaged gradient until the norm of the center of mass reaches $\sqrt N$. 
Since the initial condition for the dynamics of the center of mass is the null vector, one obtains that at the end of the first regime the center of mass position equals 
$$
\bx^{\rm CM}_{I}=-\sqrt{N}\frac{\bg}{||\bg||_2} \ .
$$
The second regime corresponds to gradient descent on the sphere with energy $H$ starting from $\bx^{\rm CM}_{I}$. \\
It is easy to check that for $N$ large the leading contribution to $g_i$
is given by terms in which the indices $i_2,\cdots,i_k$
are grouped in $(k-1)/2$ distinct pairs of the same index. In these cases $\mathbb {E}[x_{i_2}\dots x_{i_k}]$ is simply equal to one. For example, for $k=3$, one obtains: 
$$
g_i=-\frac 1 N\sum_{j}W_{i,j,j}  -  \frac{\lambda}{2N^2} v_i \left(\sum_j v_j^2\right)
=-\frac 1 N\sum_{j}W_{i,j,j}  -  \frac{\lambda}{2N} v_i \ .
$$
The first contribution to $\bg$, corresponding to $\bW$, is a random Gaussian vector with norm scaling as $N^{\frac{3-k}{4}}$ and the second is a vector in the direction of the signal, $v_i$, of norm scaling as $\lambda N^{\frac{2-k}{2}}$. If the second term is the largest, i.e. for $\lambda$ growing faster than $N^{\frac{k-1}{4}}$, a finite overlap with the signal, $m_I=(\bx^{\rm CM}_{I},\bv)/N$ is already obtained at the end of the first regime. 
See the SM for a detailed derivation of the results. 
In the following we 
focus on the more challenging SNR regime, 
$N^{\frac{k-3}{4}}\ll\lambda\ll N^{\frac{k-1}{4}}$, where
the first term has the largest norm and the overlap with the signal at the end of the first dynamical regime is approximately equal to
$$
m_I\approx\frac{1}{(k-1)!} \lambda N^{\frac{1-k}{4}} \ .
$$
How large this value of $m_I$ has to be to guarantee recovery using gradient descent in the second dynamical regime?  
The answer to this question comes from the analysis of the number of spurious minima of $H$ for configurations with overlap larger or equal to $m_I$. 
The results of \cite{BeMeMN17,RoBeBC19} obtained by the Kac-Rice method, imply that
if $\lambda m^{k-2}_I>C_k$ ($C_k$ does not scale with $N$ and is computed in the SM) then such number is not exponentially large in $N$, i.e. the initial condition for the gradient descent dynamics lies in the "easy" part of the configuration space where spurious minima that can trap the dynamics do not proliferate. This is the {\it crucial criterion} that guarantees recovery by gradient descent dynamics.\\
Let us first show that this criterion allows to recover the results for the GD algorithm. 
The initial condition for GD is a vector drawn uniformly at random on the sphere, which has typically an overlap with the signal of the order of $1/\sqrt N$.
Thus, the previous criterion requires $\lambda$ scaling as $N^{\frac{k-2}{2}}$ for gradient descent to recover the signal, which is indeed the threshold  conjectured\footnote{It was shown rigorously that $\lambda$ scaling as $N^{\frac{k-2}{2}+\frac 1 6}$ is a sufficient condition for GD initialized from a random uniform initial condition to recover the signal. As argued in \cite{BaGhJa18}, it should be possible to obtain a tighter bound and remove the $1/6$ factor by generalizing the proof  of \cite{BaGhJa18}.} in \cite{BaGhJa18} and heuristically re-derived in more details in SM1.  This is also the scaling of algorithms such as approximate message passing and Langevin dynamics \cite{MonRic14,BaGhJa18}.
SiAGD instead provides for gradient descent in the second dynamical regime an initial condition which has 
an overlap $m_I$ possibly larger than $1/\sqrt{N}$. Imposing that $\lambda m^{k-2}_I>C_k$ allows us to find the algorithmic threshold for SiAGD: 
$$
\lambda>C'_k N^{\frac{k-2}{4}}
$$
where $C'_k$ is an  $N$-independent constant that can be straightforwardly related to $C_k$. Using this scaling one finds that $m_I$  is at least of order $N^{-\frac 1 4}$.\\
We have therefore obtained two main results: we have shown that a simplified version of iAGD allows to match the performance of the best known algorithms, which is $\lambda\sim N^{\frac{k-2}{4}}$ \cite{MonRic14,HoShSt15,AnDeGM16}, 
and we have derived such an optimal algorithmic transition directly resorting to the statistical properties of the landscape.
Both results will be tested and confirmed numerically in the next section. \\
Finally, we notice that the second regime of SiAGD shares similarities with the homotopy-based algorithm studied in \cite{AnDeGM16}: they both used the same initial condition (i.e.\ what is reached at the end of the first stage of dynamics of SiAGD), but the latter consists in gradient descent with $\eta=\infty$, and this is less efficient than AGD (see discussion in SM8).

\subsection{Case II: k even}
For even values of $k$, the initial value of the average gradient is exactly zero since $\mathbb {E}[x_{i_2}\dots x_{i_k}]=0$. In this case, as discussed previously, one has to focus on the averaged Hessian, which at the initial condition of the iAGD algorithm reads: 
$$
\mathcal{H}_{ij}=-\frac{1}{N^{\frac{(k-1)}{2}}}\sum_{i_3\le\dots\le i_k}W_{i,j,i_3\dots,i_k} \mathbb {E}[x_{i_3}\dots x_{i_k}] -  \frac{\lambda\; v_i v_j}{(k-2)!N^{k-1}} \sum_{i_3,\cdots,i_k}
v_{i_3}\cdots v_{i_k}\mathbb {E}[x_{i_3}\dots x_{i_k}]
$$
The leading contribution to $\mathcal{H}_{ij}$ is given by terms in which the indices $i_3,\cdots,i_k$
are grouped in distinct pairs. In this case 
the average $\mathbb {E}[x_{i_3}\dots x_{i_k}]$ is simply equal to one. For example, for $k=4$, one obtains: 
$$
\mathcal{H}_{ij}=-\frac{1} {N^{3/2}}\sum_{k}W_{i,j,k,k}  -  \frac{\lambda}{2N^3} v_i v_j \left(\sum_k v_k^2\right)
=-\frac{1} {N^{3/2}}\sum_{k}W_{i,j,k,k}  -  \frac{\lambda}{2N^2} v_i v_j
$$
The first term of $\mathcal{H}$ is a random matrix belonging to the Gaussian Orthogonal Ensemble \cite{book}, whereas the second term is a rank one perturbation 
proportional to the projector in the direction of the signal. 
Such random matrices display an interesting phenomenon called BBP transition (Ben Arous, Baik, Pech\'e \cite{EdwJon76,BaBePe05}): given a symmetric matrix with random elements extracted from a normal distribution $\mathcal{N}(0,1/N)$ perturbed by a rank one matrix $-\alpha v_iv_j/N$ with $||\bv||_2=N$, in the large $N$ limit there exists a  finite $\alpha_{\rm BBP}=1$ such that for for $\alpha>\alpha_{\rm BBP}$ the eigenvector associated to the smallest eigenvalue of the matrix has a finite overlap with $\bv$.
By taking into account the specific scaling with $N$ of the two terms in the Hessian we get that at large $N$ for $\lambda>(k-2)!\alpha_{\rm BBP}N^{\frac{k-2}{4}}$ the eigenvector corresponding to the smallest eigenvalue of the Hessian has a finite overlap with the signal.
In consequence, for SNR above $N^{(k-2)/4}$, in the
even $k$ case, the information about the signal is present in the initial averaged Hessian: already at the beginning of the dynamics, by averaging over different replicas, a downward direction towards the signal emerges.   
At variance with the $k$ odd case, a simplified iAGD algorithm that consists in moving the center or mass in the direction of the eigenvector associated to the smallest eigenvalue of the initial averaged Hessian until hitting the sphere with radius $\sqrt N$ is already enough to obtain the best algorithmic performance. For even values of $k$, the second regime of SiAGD, corresponding to gradient descent on the sphere, is not even needed to obtain a finite overlap. \\
It is interesting to contrast the result above with the one for the Hessian obtained for a random vector drawn uniformly on the sphere, which is a typical initial condition for the GD algorithm. Repeating the previous analysis, one finds a similar result---a GOE matrix perturbed by a rank one perturbation in the direction of the signal---but now the BBP transition takes place for $\lambda>(k-2)!N^{\frac{k-2}{2}}$, which is indeed the conjectured scaling to recover the signal by gradient descent \cite{BaGhJa18}.\\
The analysis performed above can be repeated for a finite number of replicas, hence bridging the gap between the performance of the GD and SiAGD algorithm. 
For finite $R$ one finds that the algorithmic transition is at $\lambda(R)\sim N^{(k-2)/2}R^{-0.5(k-2)/(k-1)}$ (see SM3). The use of $R>1$ different initial configurations helps reducing the algorithmic gap: the larger is $R$ the smaller the algorithmic threshold is. As explained in the SM, the smoothing of the landscape using different replicas becomes ineffective when $R\gg R_{\rm opt}\sim N^{(k-1)/2}$. However, for these values of $R$ one has already reached the regime studied above.\\
In summary, in the odd and even $k$ cases, we find that the analysis of the "bare" landscape naturally leads to the scaling of the algorithmic threshold as $N^{\frac{k-2}{2}}$ whereas the analysis performed using many replicas allow to substantially averaging out the noise and to match the best scaling currently known, which is  $N^{\frac{k-2}{4}}$.\\
We have found that the $k$-even case is simpler than the $k$-odd one; this finding emerges also  from the previous literature (more involved methods were used to obtain the scaling $N^{\frac{k-2}{4}}$ for odd values of $k$), but was not explained. Our landscape based analysis offers a simple reason for it.     


\section{Numerical results}

In this section we present the results of our numerical tests, which are limited to the $k=3$ case because the memory requirements scale like $N^k$ and thus for larger values of $k$ one is limited to very small values of $N$.
The aim of this section is twofold: on the one hand we want to identify the algorithmic thresholds for both the full and simplified versions of iAGD, on the other hand we wish to directly test the connection between iAGD and SiAGD performance and the properties of the energy landscape. Numerical results for AGD with finite $R$ and their comparison with iAGD are reported in SM7. As discussed in the previous section, it was shown that there exist no spurious minima \cite{BeMeMN17} such that its overlap with the signal satisfies $\lambda m^{k-2}>C_k$ (for $k=3$ one finds $C_3\simeq 0.425815$, see SM for further details). In the following we are going to show numerically that such condition is directly related to the algorithmic threshold of iAGD. The results we present are obtained for runs of iAGD and SiAGD on problems of sizes $N=30, 100, 300, 1000, 2000$. They are then averaged over a number $M$ of different disorder realizations such that $N M=1.2\cdot 10^5$.

\begin{figure}
  \centering
  \includegraphics[width=0.48\textwidth]{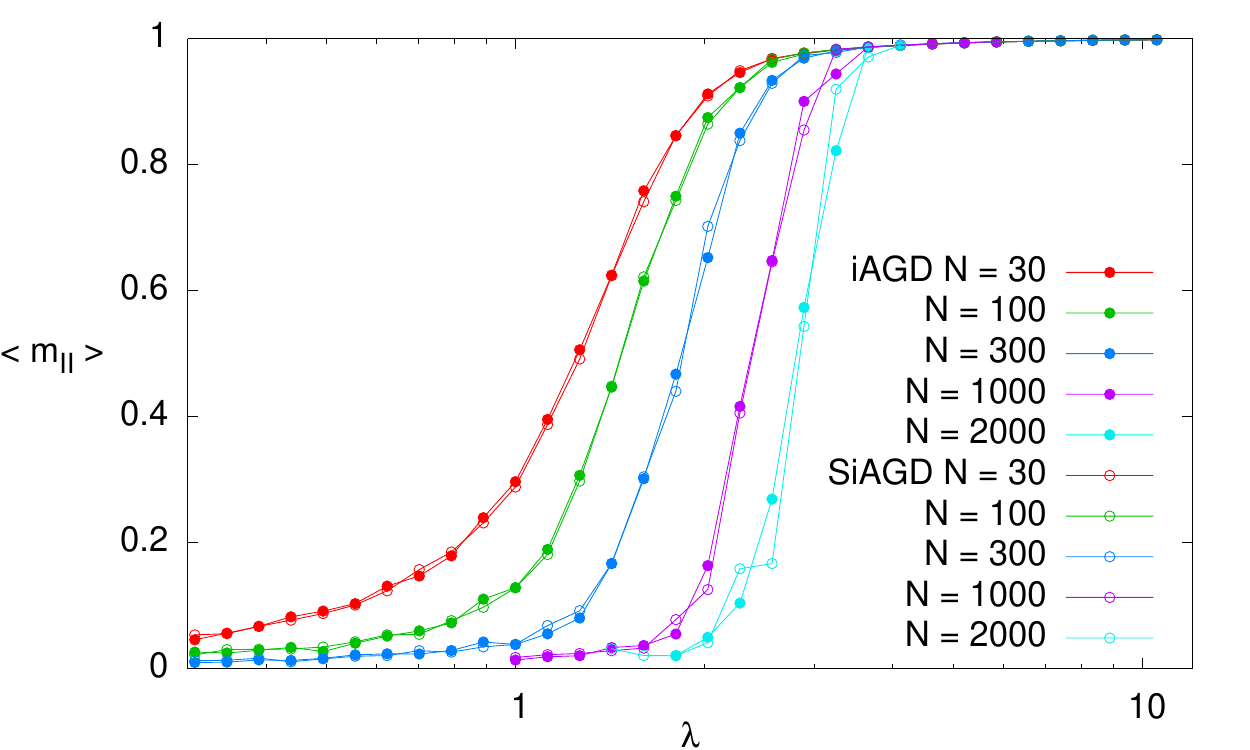}
  \hfill
  \includegraphics[width=0.48\textwidth]{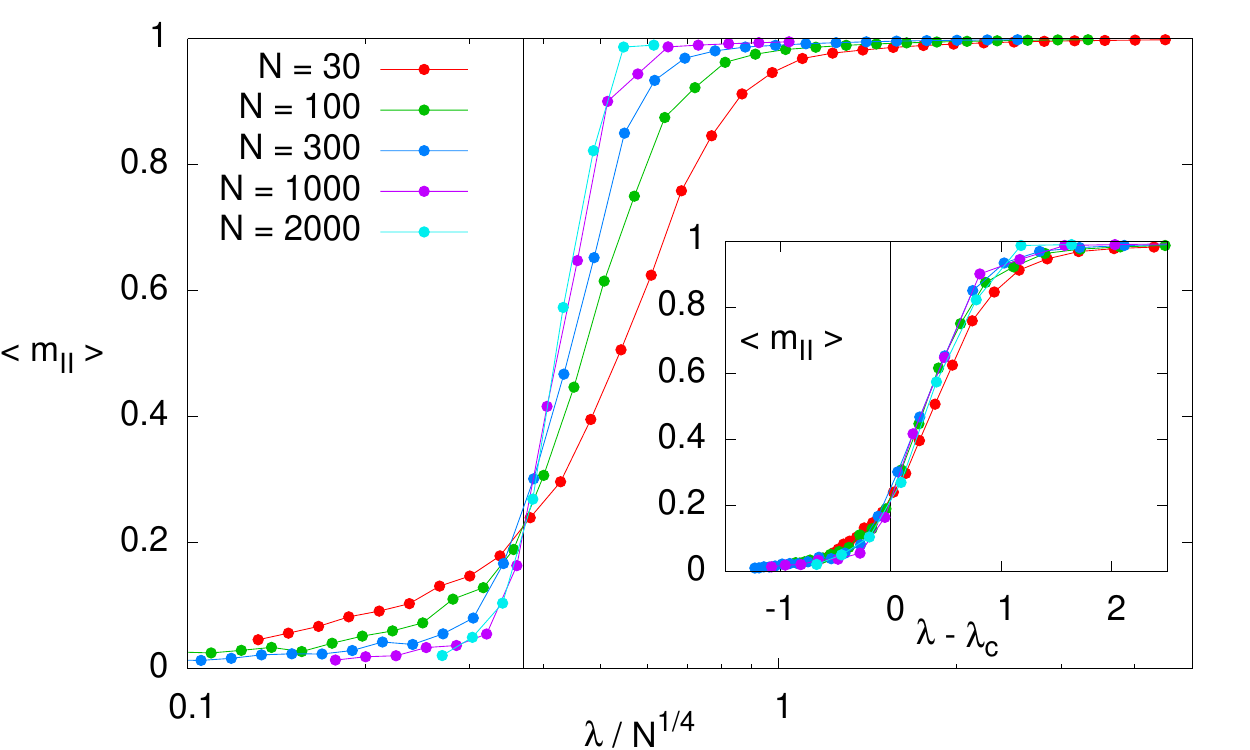}
    \caption{Left: iAGD and SiAGD achieve the same accuracy detecting the signal in tensor PCA with $k=3$. Right: their algorithmic threshold scales as $\lambda_c \simeq 0.37 N^{1/4}$ (only data for iAGD are shown). Inset: the final overlap with the signal mostly depends on $\lambda-\lambda_c$.
  \label{fig12}}
\end{figure}
\begin{itemize}
\item{\it Algorithmic threshold and threshold phenomenon}. In Figure~\ref{fig12} (left panel) we show the mean overlap with the signal, $m_{II}$, achieved at the end of the algorithm (either for iAGD or SiAGD) as a function of the signal to noise ratio $\lambda$. In the right panel we show that a threshold phenomenon (a phase transition) is taking place in the large $N$ limit on the scale $\lambda\sim N^{(k-2)/4} = N^{1/4}$.
It is worth noticing, as shown in the left panel, that both versions of the algorithm, iAGD and SiAGD, do achieve the same final mean overlap with the signal. For this reason in the right panel we have re-scaled only the data obtained via iAGD. In the right panel we also mark with a vertical line our best estimation for the critical threshold $\lambda_c \simeq 0.37 N^{1/4}$. Finally, the inset shows the same results plotted as a function of $\lambda-0.37 N^{1/4}$. This highlights that the size of the critical window around the algorithmic threshold $\lambda_c \simeq 0.37 N^{1/4}$ is almost $N$ independent. 

\item {\it Comparison between iAGD and SiAGD}. Although the final overlap achieved by the two versions of the algorithm is the same, the dynamics followed by the algorithms in the first regime is very different (see previous section for the distinction of two regimes in the dynamics). While in the SiAGD algorithm the center of mass takes a straight path to the surface of the sphere of radius $\sqrt{N}$, in the iAGD algorithm the center of mass moves according to the mean gradient at each time and thus follows a curved trajectory determined by the landscape. A priori it is unclear which dynamics is better; we offer an insight by measuring the evolution of the center of mass during and at the end of the first regime.

\begin{figure}
  \centering
  \includegraphics[width=0.48\textwidth]{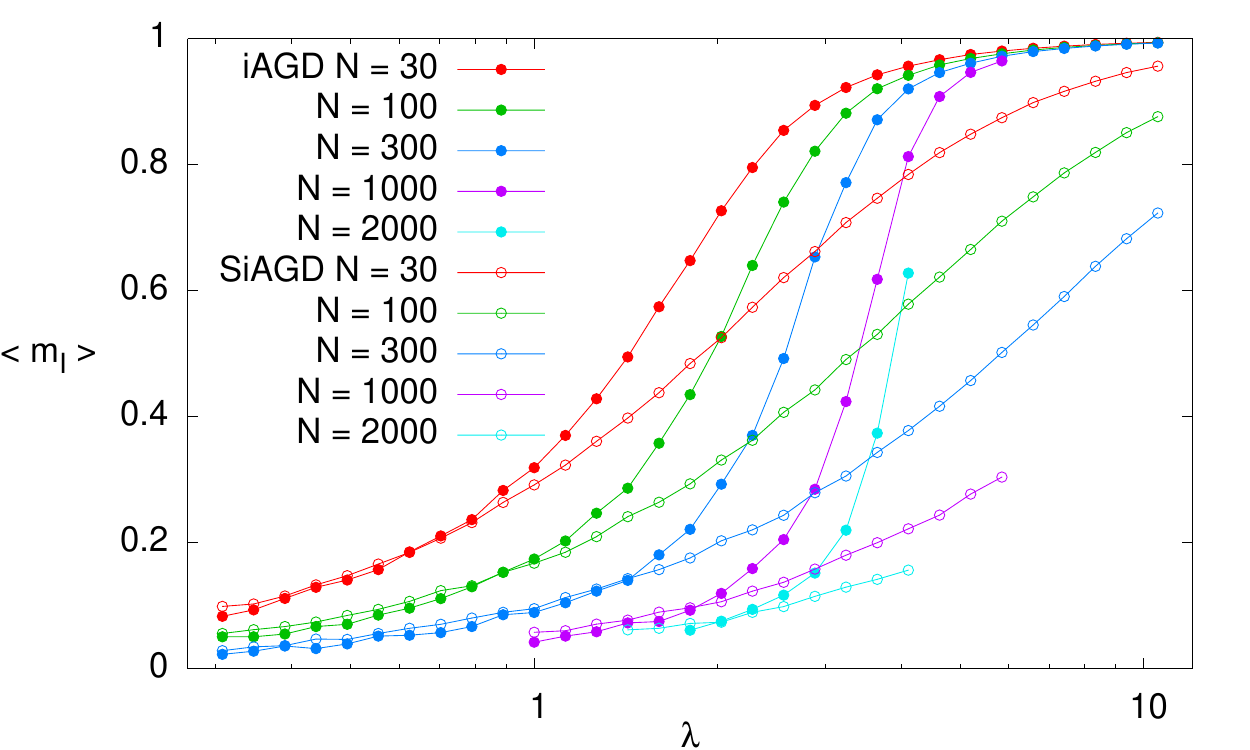}
  \hfill
  \includegraphics[width=0.48\textwidth]{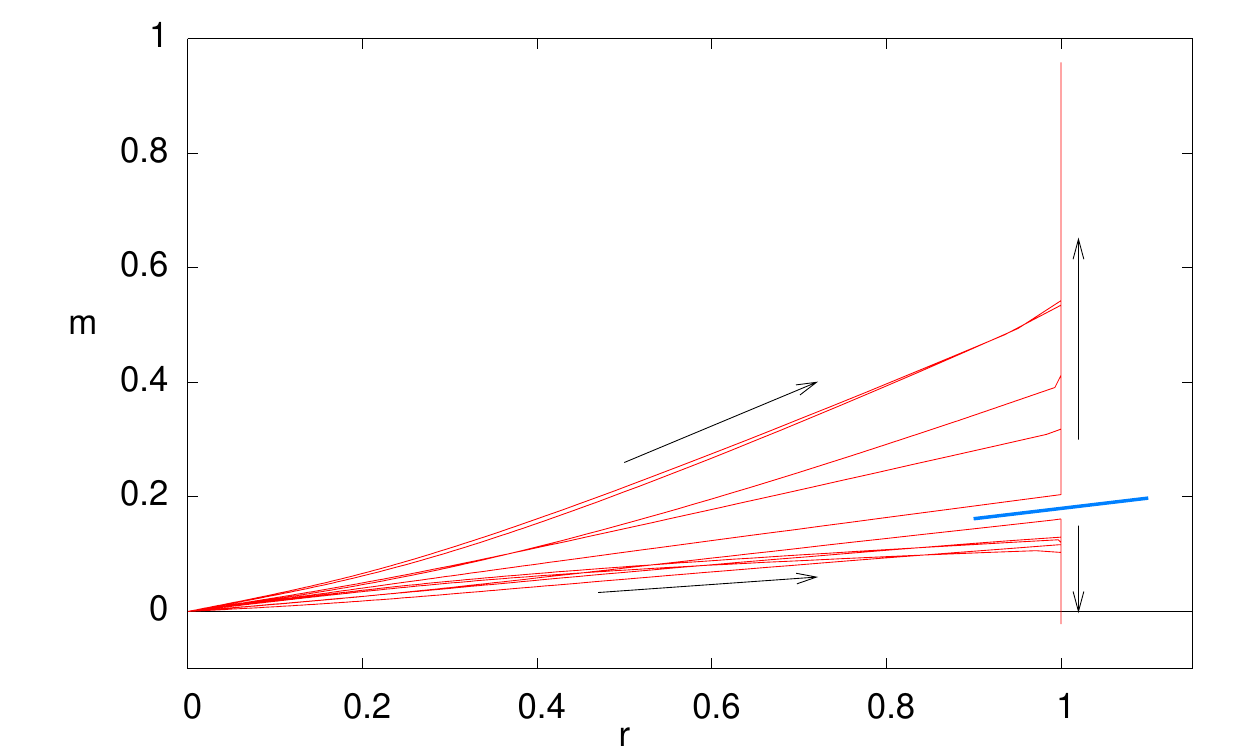}
  \caption{Left: at the end of the first dynamical regime iAGD achieves an overlap with the signal larger or equal to the one achieved by SiAGD. Right: A schematic picture of the trajectories followed by iAGD, represented by $\rCM=||\bxCM||_2/\sqrt{N}$ and the overlap $m$ with the signal. \label{fig56}}
\end{figure}

In the left panel of Figure \ref{fig56} we report the mean overlap $\langle m_I\rangle$ achieved at the end of the first phase by the iAGD and SiAGD algorithms. We clearly see that the dynamics followed by the iAGD algorithm reaches a larger overlap. Therefore a natural question arises: how can SiAGD achieve the same accuracy in detection than iAGD although it starts from a lower value of $m_I$? While trying to answer this question, we notice an important difference between the two dynamics in the first phase: although both depend on the landscape, they feel the landscape in a quite different way.
In the SiAGD algorithm the mean gradient is computed only once at the beginning. Then a straight path is followed until the center of mass hits the sphere. In this sense the algorithm in its first regime should be considered as a strongly out of equilibrium process that feels little of the original landscape and thus ends on a point on the sphere whose energy has not been optimized.
SiAGD then secures its own connection to the landscape only in the second regime, where it continues with usual gradient descent that starts from this high energy configuration.\\
iAGD starts in a similar way computing the mean gradient when the centre of mass is close to the origin. 
At this initial stage, the averaging process reaches its highest efficiency in ironing out the landscape as the replicas are completely uncorrelated. The gradient on the center of mass is much less affected by noise with respect to the one of single replicas. 
However, as soon as the centre of mass starts to approach the sphere of radius $\sqrt{N}$ the cloud of replicas shrinks, thus sampling a progressively smaller region of the landscape, until the mean gradient converges continuously to the standard gradient.
Thus we expect iAGD to reach a point on the sphere of lower energy than SiAGD.
This is explicitly shown in SM6.
In summary iAGD and SiAGD algorithms reach the same accuracy in signal detection, although they land on the sphere on very different points, with iAGD reaching larger overlaps and lower energies.

\item {\it Landscape and dynamics}. In the right panel of Figure \ref{fig56} we show the trajectories followed by the center of mass during the execution of the iAGD algorithm solving 10 problems of size $N=300$ with $\lambda=2$: we plot the overlap of the center of mass with the signal $m=(\bxCM,\bv)/N$ versus the normalized norm of the center of mass $\rCM=||\bxCM||_2/\sqrt{N}$.
Recall that when $\rCM=1$ iAGD reduces to standard GD.
Observing the plot it should be clear that there is a threshold value for the overlap on the sphere (marked by a thick blue line) such that when the algorithm hits the sphere above (below) that threshold value, then GD is able (not able) to recover the signal.
\begin{figure}
  \centering
  \includegraphics[width=0.48\textwidth]{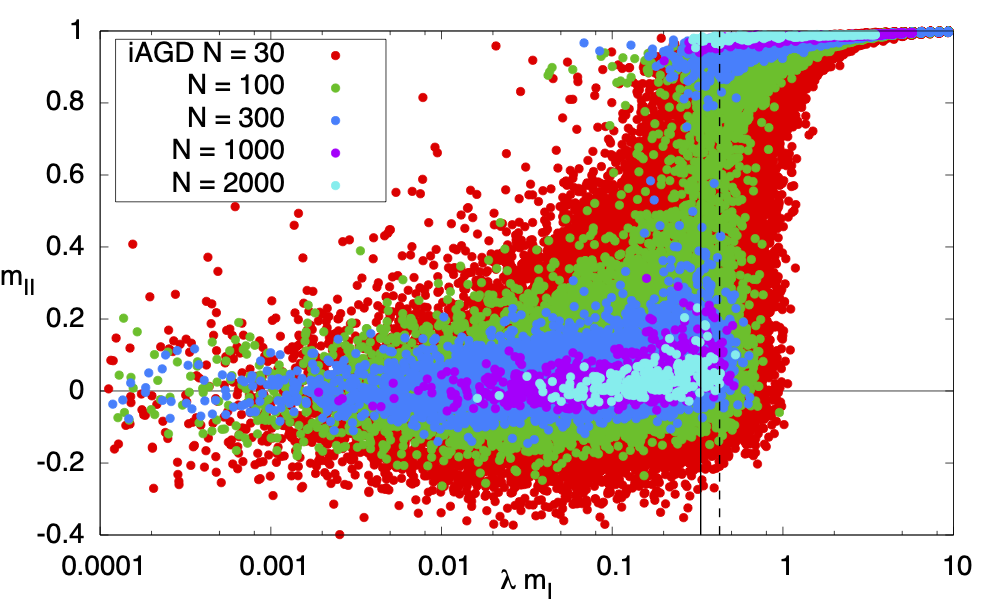}
  \hfill
  \includegraphics[width=0.48\textwidth]{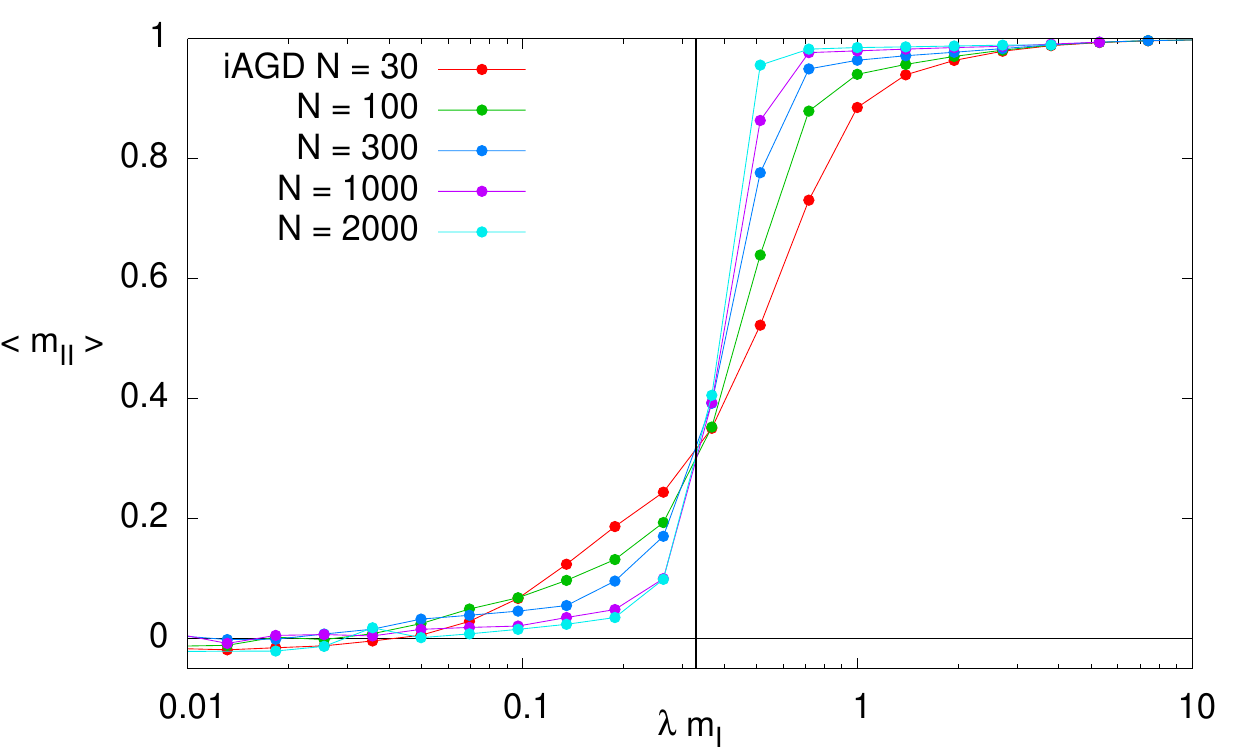}
  \caption{Signal detection is possible if $\lambda m_I$ is larger than the threshold value reported with a full vertical line, estimated from averaged data in the right panel. Complexity is null on the right of the dashed line. \label{fig78}}
\end{figure}
Moreover we notice that the trajectories of the runs that eventually detect the signal tend to bend upwards already in the first dynamical regime.\\
To better illustrate the threshold phenomenon in $m_I$ we show in the left panel of Figure \ref{fig78} a scatter plot of the final overlap $m_{II}$ versus $\lambda m_I$. 
Clouds of points have different sizes for two reasons: for the smaller problems we have studied more samples and finite size effects tend to disperse the points more for smaller sizes. 
We clearly see that for large enough $N$ the data points form two different and well separated clouds: the lower one corresponds to samples where iAGD has been unable to detect the signal, while the upper one corresponds to samples where signal detection was achieved. The choice of using a scaled overlap $\lambda m_I$ for the abscissa is dictated by the observation that the complexity of local minima depends only on the variable $\lambda m_I^{k-2}=\lambda m_I$ (for $k=3$) in the large $N$ limit and it is null with high probability for $\lambda m_I > C_3=0.425815$ (marked by a dashed vertical line in the plot). The full vertical line marks the location of the threshold estimated from the data shown in the right panel of Figure \ref{fig78}: in the large $N$ limit if iAGD reaches an overlap satisfying $\lambda m_I \gtrsim 0.33$ then it detects the signal with high probability. 
We have thus found that the numerically estimated threshold is slightly lower than the one where spurious minima disappear. This can be due to multiple reasons: first the result \cite{BeMeMN17} used to estimated the number of minima only provides an upper bound, a quenched Kac-Rice computation \cite{RoBeBC19} would be needed to obtain the exact value. 
Second, very recently it has been shown that landscape-based algorithm, such as GD, can succeed even in presence of spurious minima \cite{MaKrUZ19}.
Moreover it has been also shown that the minima where these landscape-based dynamics end may depend on the starting energy and the most attracting minima are not the most numerous ones \cite{FoFrRT19}.
The inspection of this issue in further details is left for future work.
\end{itemize}
\section{Conclusions and discussion}
We have proposed a new algorithm which is a generalization of gradient descent and uses the idea that by averaging the gradient of uncorrelated copies of the system one can substantially reduce the roughness of the landscape.  
One of its main advantages is its generality; in fact, AGD
can be straightforwardly and directly applied to many hard inference problems without any prior knowledge. \\
The spiked tensor problem has provided the perfect ground-test to study its performance, showing that it is slightly better (in the prefactor) than the state-of-the-art algorithms and much better (in the $N$ scaling) than other algorithms based on the landscape.\\
It is worth discussing the superiority of AGD to AMP. The latter is often the provably best algorithm for models where variables interact in a dense and asymptotically very weak way. However the spiked tensor is one of those problems where AMP is sub-optimal therefore it come as no surprise that other algorithms can do better. 
An important message from our work is that the information locally available to AMP is much smaller than the one that can be collected with many uncorrelated replicas allowing AGD to reach much better performances.
Another way of understanding the deep difference between AGD and AMP-like algorithms is to consider the way the elements of the tensor are used by these different algorithms: in AMP all elements are used with a similar weight, while our algorithm gives much more weight to elements having pairs of identical indices. The two  algorithms are extracting different information from the same tensor.\\
We have studied different versions of the algorithm --- AGD, iAGD, SiAGD --- because on the one hand the versions with infinite $R$ can be solved analytically for the spiked tensor problem, on the other hand the version that presents the best performances is the one with finite $R$ (see SM7) and shows that AGD has the potential to achieve unprecedented results already when working with a limited number of replicas.\\
Not only AGD outperforms the best available algorithm for signal recovery in the spiked tensor problem (comparison with Homotopy is shown in SM8) but we believe it can be straightforwardly extended to more general problems in Machine Learning. For example we expect that problems where in general a low-rank signal, say a $P$-dimensional signal, is hidden by additive noise (in this manuscript we considered the $P=1$ case) can be solved by AGD as the average gradient or the average lowest Hessian eigenvector would point towards the $P$-dimensional subspace containing the signal.
Additional work on the application of AGD to problems of this kind is required to substantiate these claims.

\subsubsection*{Acknowledgments}
We thank for interesting and very useful discussions Marco Baity-Jesi, Gerard Ben Arous, Aukosh Jagannath, Florent Krzakala, Marc M\'ezard, Andrea Montanari, Lenka Zdeborova.
This work has been conceived and mainly developed at the Kavli Institute for Theoretical Physics within the program entitled ``The Rough High-Dimensional Landscape Problem'', as such this research was supported in part by the National Science Foundation under Grant No. NSF PHY-1748958. We also acknowledge support by the Simons Foundation collaboration Cracking the Glass Problem (No. 454935 to G. Biroli and No. 454949 to G. Parisi).
\vspace{1cm}
\bibliographystyle{unsrt}
\bibliography{bib}

\newpage

\mbox{ }

\section*{\Large Supplementary Material}

\vspace{5mm}

\section*{SM1: Gradient Descent}
We provide here more detailed arguments to derive the algorithmic transition of gradient based methods applied to the spiked tensor problem.
The aim of the algorithm is to exit as soon as possible the region on the equator populated by uninformative spurious minima which may trap a gradient descent algorithm.
As discussed in the main text section, it was shown that there exist no spurious minima \cite{BeMeMN17} having an overlap $m$ with the signal which satisfies $\lambda m^{k-2}>C_k$ (see also SM5).
 The region closer to the signal is expected to be free from minima although full of all sorts of other stationary points, {\it i.e.} saddles. \\
Usual gradient descent algorithm (GD) will naturally be able to retrieve the signal starting from a random initial condition with $m\sim 1/\sqrt{N}$ as soon as the SNR is $\lambda\gg\lambda_{\rm GD}\sim N^{(k-2)/2}$ so that $\lambda m^{k-2}\gg 1$. 
More specifically we can reobtain the same result looking at the initial gradient, which reads
\begin{equation}
    g_i^{\rm GD}=\nabla_i H \propto -N^{\frac{-(k-1)}{2}}\sum_{i_2,\dots,i_k} W_{i,i_2,\dots,i_k} x_{i_2}\dots x_{i_k}
    - \lambda v_i m^{k-1} \ .
\end{equation}
This vector has norm $|{\bf g}_{\rm GD}|\sim N^{(2-k)/2}\sqrt{(N^{k-1}+\lambda^2)}$ and, when normalized as to point on the surface of the sphere $S^{N-1}(0,\sqrt{N})$, the following projection on the signal $\bf v$
\begin{equation}
    m_{\rm GD}=-\frac{{\bf v \cdot g}_{\rm GD}}{\sqrt{N}|{\bf g}_{\rm GD}|}
    \propto \frac{(N^{(k-2)/2}+\lambda)}{\sqrt{(N^{k-1}+\lambda^2)}} \ .
\end{equation}
We distinguish three regimes in terms of the SNR.
The first where it is $\lambda\ll N^{(k-2)/2}$, $m_{\rm GD}\sim 1/\sqrt{N}$, and $\lambda m_{\rm GD}^{k-2}\ll 1$.
In the second we have $N^{(k-2)/2}\ll \lambda\ll N^{(k-1)/2}$, therefore $m_{\rm GD}\sim \lambda/N^{(k-1)/2}\ll 1$, but $\lambda m_{\rm GD}^{k-2}\gg 1$.
In the third, when $\lambda\gg N^{(k-1)/2}$, we immediately have $m_{\rm GD}= 1$.
The interesting algorithmic threshold is therefore at $\lambda_{\rm GD}\sim N^{(k-2)/2}$ where the information contained in the gradient is enough to escape from the region full of minima. At this point subsequent steps of gradient descent are needed for the reconstruction of the signal, but the success is granted. Starting from $\lambda\sim N^{(k-1)/2}$ the recovery of the signal is instead obtained at the first step.

\section*{SM2: Simplified Averaged Gradient Descent}
Let's apply the same reasoning to the case of Simplified Averaged Gradient Descent (SAGD), with $R$ initial copies of the system all at $m_{\alpha}\sim 1/\sqrt{N}$.
The gradient at the first step now reads
\begin{align}
    g_i^{\rm SAGD}&=\frac1R \sum_{\alpha}\nabla_i H|_{{\bf x}_{\alpha}}\propto\nonumber\\
    &\propto -\frac{N^{\frac{-(k-1)}{2}}}{R}\sum_{\alpha}\sum_{i_2,\dots,i_k} W_{i,i_2,\dots,i_k} x^{\alpha}_{i_2}\dots x^{\alpha}_{i_k}
    - \frac{\lambda}{R} v_i \sum_{\alpha}m_{\alpha}^{k-1} \ ,
\end{align}
with norm $|{\bf g}_{\rm SAGD}|\sim N^{(2-k)/2}R^{-1/2}\sqrt{(N^{k-1}+R\lambda^2)}$ for odd $k$ and projection on ${\bf v}$, after normalization on the sphere, 
\begin{equation}
    m_{\rm SAGD}
    \propto \frac{(N^{(k-2)/2}+\sqrt{R}\lambda)}{\sqrt{(N^{k-1}+R\lambda^2)}} \ .
\end{equation}
Again we distinguish three regimes.
The first where it is $\lambda\ll N^{(k-2)/2}/\sqrt{R}$, $m_{\rm SAGD}\sim 1/\sqrt{N}$, and $\lambda m_{\rm SAGD}^{k-2}\ll 1/\sqrt{R} <1$.
In the second regime we have $N^{(k-2)/2}/\sqrt{R}\ll \lambda\ll N^{(k-1)/2}/\sqrt{R}$, therefore $m_{\rm SAGD}\sim \sqrt{R}\lambda/N^{(k-1)/2}\ll 1$, and $\lambda m_{\rm SAGD}^{k-2}\sim \lambda^{k-1}R^{(k-2)/2}N^{-(k-1)(k-2)/2}$, which implies a $$\lambda_{\rm SAGD}\sim N^{(k-2)/2}R^{-0.5(k-2)/(k-1)} \ .$$
Note that the algorithmic transition lies in this second SNR regime in all the interesting cases. Indeed for $R>1$ it is always $\lambda_{\rm SAGD}> N^{(k-2)/2}/\sqrt{R}$, while $\lambda_{\rm SAGD}\ll N^{(k-1)/2}/\sqrt{R}$ holds only if $\sqrt{R}/N^{(k-1)/2}\ll 1$, hence $R\ll N^{(k-1)}$, which is always satisfied by the optimal number of replicas $R_{\rm opt}$
that are needed to achieve the best algorithmic performances of AGD as we will see in section SM3 and SM4.
In the third regime, when $\lambda\gg N^{(k-1)/2}/\sqrt{R}$, we immediately have $m_{\rm SAGD}\sim 1$.

\section*{SM3: Optimal number of replicas}

We derive here how many replicas are needed to best iron the landscape of the spiked tensor problem, {\it i.e.} to reduce the fluctuations of the empirical average ${\bf g}_{{\rm n},R}=\sum_{\alpha}{\bf g}_{{\rm n},\alpha}/R$ of the uninformative component of the gradient ${\bf g}_{{\rm n},\alpha}$ below its population average ${\bf g}_{{\rm n}}^{\rm av}=\mathbb {E}[{\bf g}_{{\rm n},R}]$. By using the central limit theorem, it is clear that it is not useful to increase the value of $R$ above the point at which the fluctuations of ${\bf g}_{{\rm n},R}$ become smaller than its average. In order to obtain this value $R_{\rm opt}$
we evaluate the population average of each of its components as
\begin{align}
g_{{\rm n},i}^{\rm av} &= -N^{-\frac{(k-1)}{2}} \sum_{i_2\le\dots\le i_k} W_{i,i_2\dots,i_k} \mathbb {E}[x_{i_2}\dots x_{i_k}] =\nonumber\\
&= -N^{-\frac{(k-1)}{2}} (k-2)!!
\sum_{i_2\le\dots\le i_k}W_{i,i_2\dots,i_k} \delta_{i_2,i_3}\dots \delta_{i_{k-1},i_k}\;.
\end{align}
Using again the central limit theorem but now with respect to the randomness due to the choice of $\bf W$, we find that the variance of $g_{{\rm n},i}^{\rm av}$ scales like $N^{-(k-1)/2}$.
We are interested in understanding how does it compare with the population variance
\begin{equation}
\mathbb {E}[(g_{{\rm n},R,i}-g_{{\rm n},i}^{\rm av})^2]
= \mathbb {E}[g_{{\rm n},R,i}^2]-{g_{{\rm n},i}^{\rm av}}^2
\end{equation}
where 
\begin{eqnarray}
\mathbb {E}[g_{{\rm n},R,i}^2] = N^{-(k-1)}R^{-2} \hspace{-0.3cm}\sum_{\substack{i_2\le\dots\le i_k\\i'_2\le\dots\le i'_k}} \hspace{-0.3cm}W_{i,i_2\dots,i_k} 
W_{i,i'_2\dots,i'_k}
\sum_{\alpha \alpha'}
\mathbb {E}[x^{\alpha}_{i_2}\dots x^{\alpha}_{i_k}x^{\alpha'}_{i'_2}\dots x^{\alpha'}_{i'_k}] \ .
\end{eqnarray}
The dominant non zero terms in this equation are the following:
(i) if $\alpha=\alpha'$ and $(i'_2,\ldots,i'_k)$ is a permutation of $(i_2,\ldots,i_k)$ we get a contribution $N$-independent and scaling as $R^{-1}$ (we ignore $k$-dependent factors as we are mainly interested in the scaling in $N$ and $R$); (ii) if indices $(i_2,\ldots,i_k)$ are matched in pairs as well as indices $(i'_2,\ldots,i'_k)$, then the sum over $\alpha$ and $\alpha'$ cancels the $R^{-2}$ factor and we get a term identically equal to ${g_{{\rm n},i}^{\rm av}}^2$.
Therefore only the terms of the first kind are left and the population variance scales like $R^{-1}$. Of course, it is not useful to make the latter smaller than the variance of $g_{{\rm n},i}^{\rm av}$ that scales like $N^{-(k-1)/2}$, {\it i.e.} having $R>R_{\rm opt}$ with $R_{\rm opt}\sim N^{(k-1)/2}$.
In fact, a larger number of replicas would imply a larger computational effort without net advantage on the algorithmic performances. 


\section*{SM4: Best results achieved with Simplified Averaged Gradient Descent}

Putting together the results from SM3 in the discussion of SM2, we obtain that the best $m_{\rm SAGD}$ can be achieved when $R\sim R_{\rm opt}$ and it is 
\begin{equation}
    m_{\rm SAGD}(R_{\rm opt})
    \propto \frac{(N^{(k-3)/4}+\lambda)}{\sqrt{(N^{(k-1)/2}+\lambda^2)}} \ .
\end{equation}
The three SNR regimes are therefore as follows.
The first where it is $\lambda\ll N^{(k-3)/4}$, $m_{\rm SAGD}\sim 1/\sqrt{N}$, and $\lambda m_{\rm SAGD}^{k-2}\ll N^{-(k-1)/4} <1$.
In the second regime we have $N^{(k-3)/4}\ll \lambda\ll N^{(k-1)/4}$, therefore $m_{\rm SAGD}\sim \lambda/N^{(k-1)/4}\ll 1$, and $\lambda m_{\rm SAGD}^{k-2}\sim \lambda^{k-1}N^{-(k-1)(k-2)/4}$, which implies
\begin{equation}
\lambda_{\rm SAGD}\sim N^{(k-2)/4} \;,
\end{equation}
which, as expected, also coincides with the algorithmic transition for the SAGD algorithm in the limit of infinite copies that we called SiAGD and discussed in the main text.
Note finally that $R< R_{\rm opt}\sim N^{(k-1)/2}$ means that we always have $\sqrt{R}/N^{(k-1)/2}< \sqrt{R_{\rm opt}}/N^{(k-1)/2} \sim N^{-(k-1)/4}\ll 1$, which assures that the algorithmic threshold for AGD always lies in the second regime of SNR as anticipated in section SM3. Indeed even for $R=R_{\rm opt}$ it is $\lambda_{\rm SAGD}> N^{(k-3)/4}$, and $\lambda_{\rm SAGD}\ll N^{(k-1)/4}$.
The third regime, when $\lambda\gg N^{(k-1)/4}$, leads to a trivial recovery, as we commented in the main text, because it immediately gets $m_{\rm SAGD}\sim 1$.

\section*{SM5: Kac-Rice results and the criterion for the absence of spurious minima}

The  number and location of minima of $H(x)$ have been studied in \cite{BeMeMN17} . For the sake of completeness we report here the result of that work which is relevant for the purpose of understanding the behavior of GD-like algorithms. We are interested in studying the number of minima in the limit of large $\lambda$ and small overlap $m$, such that $\lambda m^{k-2}$ is constant in the large $N$ limit. Under this condition the annealed complexity, i.e.\ the normalized log of the mean number of minima, is given by
\begin{equation}
\Sigma = \frac12 \ln(k-1) +\frac2k - \frac74 + \left(1-\frac\theta4\right)\theta-\frac12\ln(\theta)
\end{equation}
where $\theta=\sqrt{2k(k-1)}\lambda m^{k-2}$. The above expression holds for $\theta>1$ and it is easy to check that it is monotonously decreasing in $\theta$ with a root in $\theta^*(k)$. For example $\theta^*(3)\simeq 1.47507$ that corresponds to $C_3=\lambda m\simeq 0.425815$, which is the value quoted in the main text.

\section*{SM6: More on numerical simulations of iAGD and SiAGD}

The tensor used in numerical experiments is obtained by symmetrizing a random tensor
\begin{equation}
    T_{i_1\ldots i_k} = \frac{\lambda}{N^{k-1}} v_{i_1}\ldots v_{i_k} + \frac{1}{N^{(k-1)/2}} \frac{1}{k!} \sum_{\text{perm.}\; i_1\ldots i_k} W_{i_1\ldots i_k} 
\end{equation}
where $W_{i_1\ldots i_k}$ are i.i.d. Gaussian random variables of zero mean and unit variance. This construction builds a tensor where the variance of elements changes depending on how many indices are equal. However this difference does not alter the results presented in the main text neither the expression for the complexity shown in SM5, since the fraction of tensor elements with a different variance is a vanishing fraction in the large $N$ limit.

The energy function we wish to minimize is
\begin{equation}
H(\mathbf{s}) = - \sum_{i_1\le i_2\le \ldots \le i_k} T_{i_1 \ldots i_k} s_{i_1} \ldots s_{i_k}
\end{equation}
under the constraint $\sum_i s_i^2=N$.

For the reader convenience we rewrite here the equations to be solved by the iAGD algorithm, {\it i.e.} in the $R\to\infty$ limit; we focus on the specific case $k=3$, which is the one we actually solved numerically. The center of mass starts from the origin $\bxCM(0)=\mathbf{0}$ and evolves according to the following differential equation
\begin{equation}
\label{eq:xCM}
\partial_t\xCM_i(t) = \left[1-r^2(t)\right] D_i + \sum_{1\le j \le k \le N} T_{ijk}\, \xCM_j(t) \xCM_k(t)
\end{equation}
with $r^2(t)=||\bxCM(t)||_2^2/N$ and $D_i=\sum_j T_{ijj}$. We solve the differential equations in Eq.~(\ref{eq:xCM}) via the Euler method with a fixed integration step $dt=0.125$ (we have checked the results do not depend on this choice). When the condition $r^2(t)\ge 1$ is met, then the algorithm continues as a standard gradient descent on the sphere of radius $\sqrt{N}$.

The equations for SiAGD are even simpler, given that the first dynamical regime consists in a single step bringing the center of mass directly on the sphere at the position determined by
\begin{equation}
    \xCM_i = \sqrt{N}\frac{D_i}{||\mathbf{D}||_2}
\end{equation}

\begin{figure}
  \centering
  \includegraphics[width=0.7\textwidth]{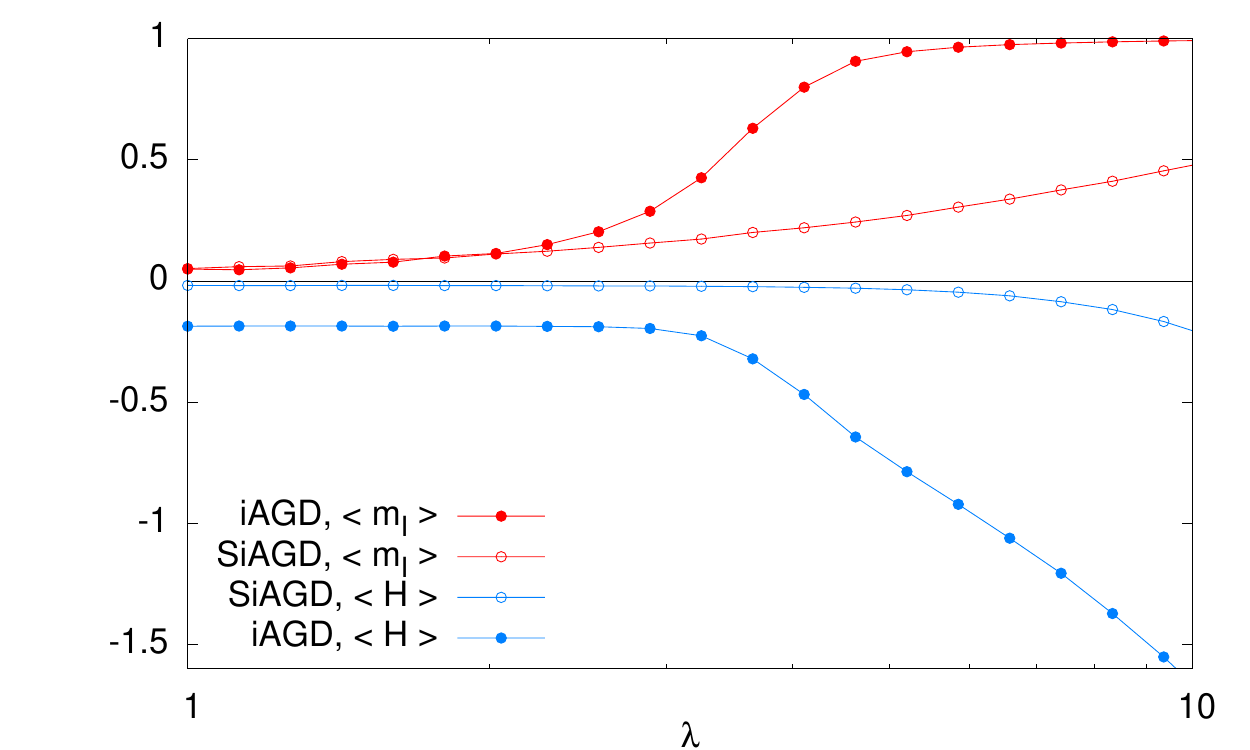}
  \caption{Mean overlap and mean energy reached by iAGD and SiAGD at the end of the first dynamical regime, when the center of mass reaches the sphere of radius $\sqrt{N}$. Data are for $N=1000$.\label{figEM}}
\end{figure}

We start showing the very different behavior of the two algorithms (iAGD and SiAGD) during the first dynamical regime.
We plot in Figure \ref{figEM} both the mean overlap $\langle m_I \rangle$ and the mean energy $\langle H \rangle/N$ of the point on the sphere reached at the end of the first dynamical regime. Data are for $N=1000$. We see not only the difference in the overlap already noticed in the main text, but also a clear difference in energy. For low values of $\lambda$, when the algorithm behavior is not strongly determined by the signal, the SiAGD algorithm reaches a point on the sphere which is random to a large extent and thus its mean energy is very close to zero. iAGD instead reaches points with a lower mean energy. Notwithstanding the very different points reached on the sphere, the final accuracy of both algorithms is very similar (as shown in the main text).

\begin{figure}
  \centering
  \includegraphics[width=0.48\textwidth]{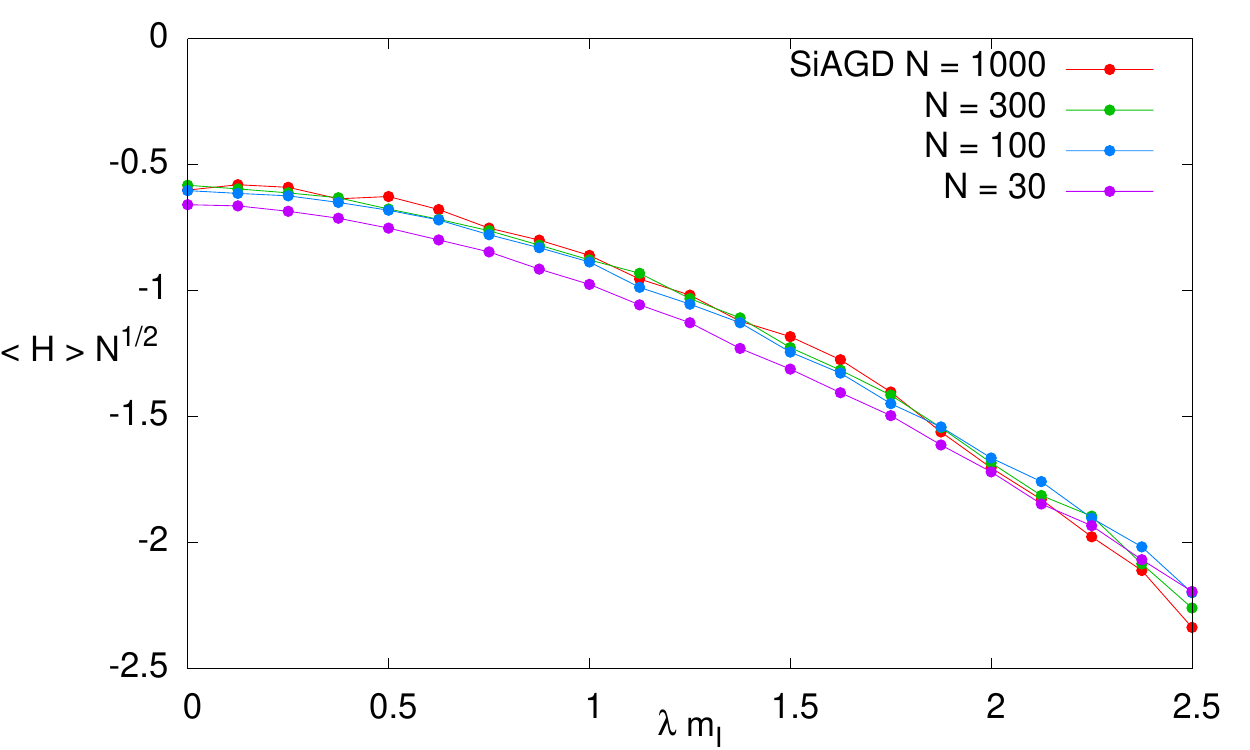}
  \hfill
  \includegraphics[width=0.48\textwidth]{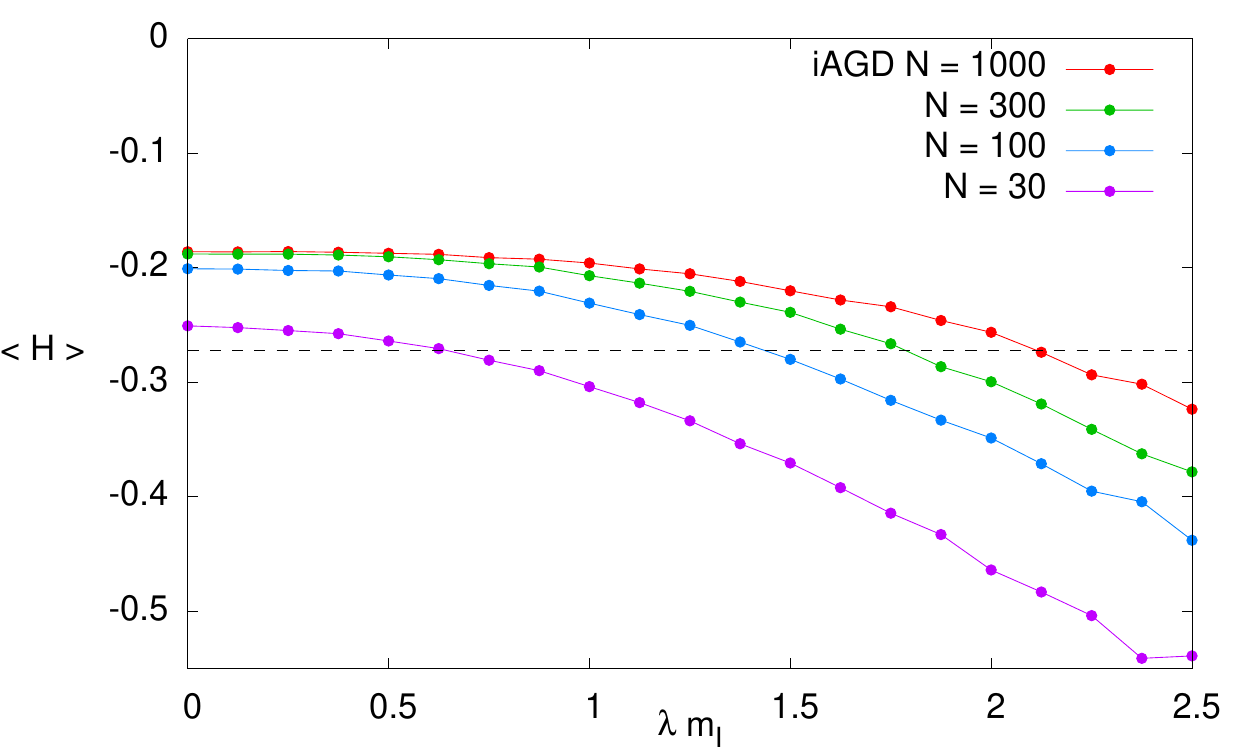}
  \caption{Mean energy at the end of the first dynamical regime, when the center of mass reaches the sphere of radius $\sqrt{N}$ for SiAGD (left panel) and iAGD (right panel). Notice the different scaling of energies: for SiAGD the mean energy has been multiplied by $\sqrt{N}$. We have used the scaling variable $\lambda m_I$ on the abscissa. The dashed horizontal line in the right panel marks the threshold energy above which there are no energy minima in the large $N$ limit. \label{figEner}}
\end{figure}

In Figure \ref{figEner} we show data for the energy reached at the end of the first regime with several values of $N$. We use the scaling variable $\lambda m_I$ for the abscissa, which allow us to average together data collected with many different values of $\lambda$. For the SiAGD algorithm the mean energy scales like $\langle H \rangle/N \sim O(N^{-1/2})$ as shown in the left panel of Figure~\ref{figEner}. For the iAGD algorithm the mean energy seems to have a well defined value close to -0.2 in the large $N$ limit. In order to make sense of this number we have reported with a dashed horizontal line the value of the threshold energy $E_{th}=-\sqrt{2/3}/3\simeq -0.272166$, above which there are no local minima uncorrelated with the signal \cite{auffinger2013random,castellani2005spin}.

The analysis of the mean overlap and mean energy at the end of the first dynamical regime suggests the following qualitative picture. Above the critical threshold (that corresponds to the scaling variable $\lambda m_I \sim 0.33$) both iAGD and SiAGD are able to move towards the signal without getting trapped by the exponentially many local minima induced by the random part of the energy function.

\begin{figure}
  \centering
  \includegraphics[width=0.48\textwidth]{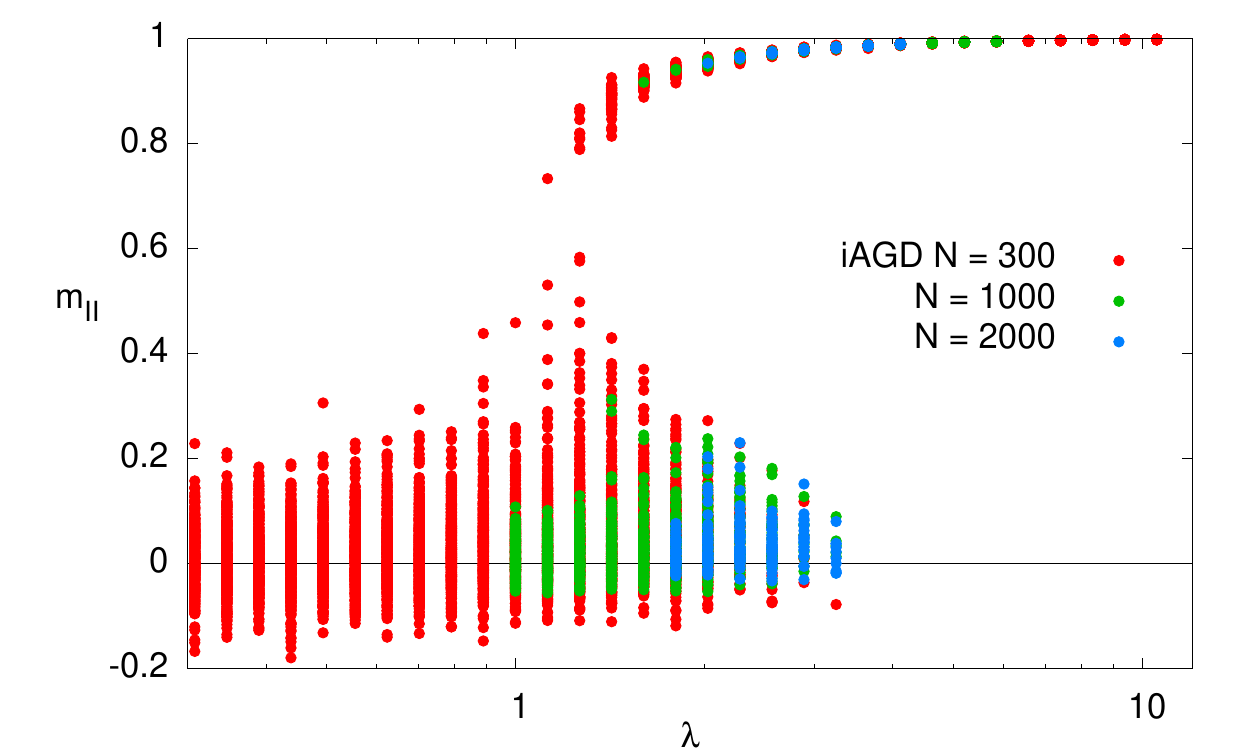}
  \hfill
  \includegraphics[width=0.48\textwidth]{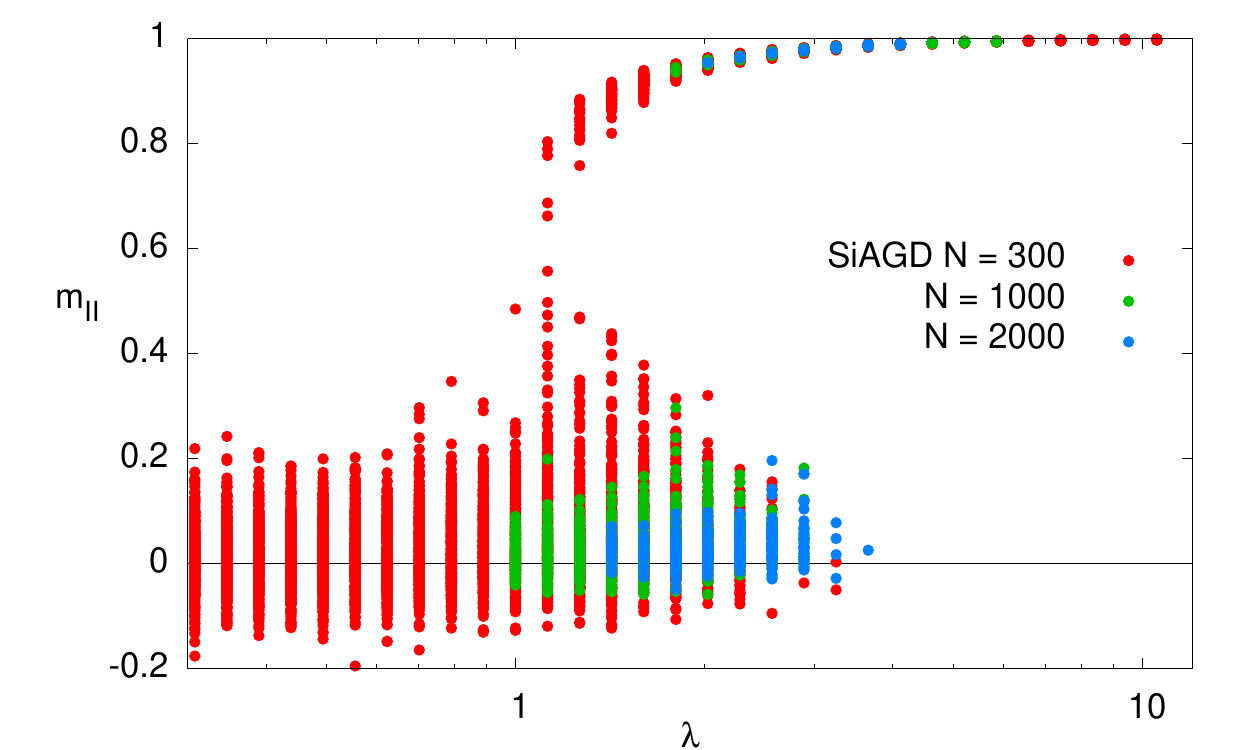}
  \caption{The phase transition leading to signal detection is discontinuous for both iAGD (left) and SiAGD (right) algorithms. \label{fig3}}
\end{figure}

We move now to discuss an aspect that we have voluntarily overlooked in the main text, that is the estimation of the statistical error on the mean overlap. The reason why we have not provided a statistical error on $\langle m_{II} \rangle$ should be clear observing data in Figure~\ref{fig3}. The overlap $m_{II}$ reached at the end of the iAGD and SiAGD algorithms shows a clear bimodal distribution close to the threshold value $\lambda \approx \lambda_c$. In such a situation the mean overlap $\langle m_{II} \rangle$ is not the most informative parameter and its statistical error is dominated by fluctuations in the fractions of points in one of the two clouds.

\begin{figure}
  \centering
  \includegraphics[width=0.48\textwidth]{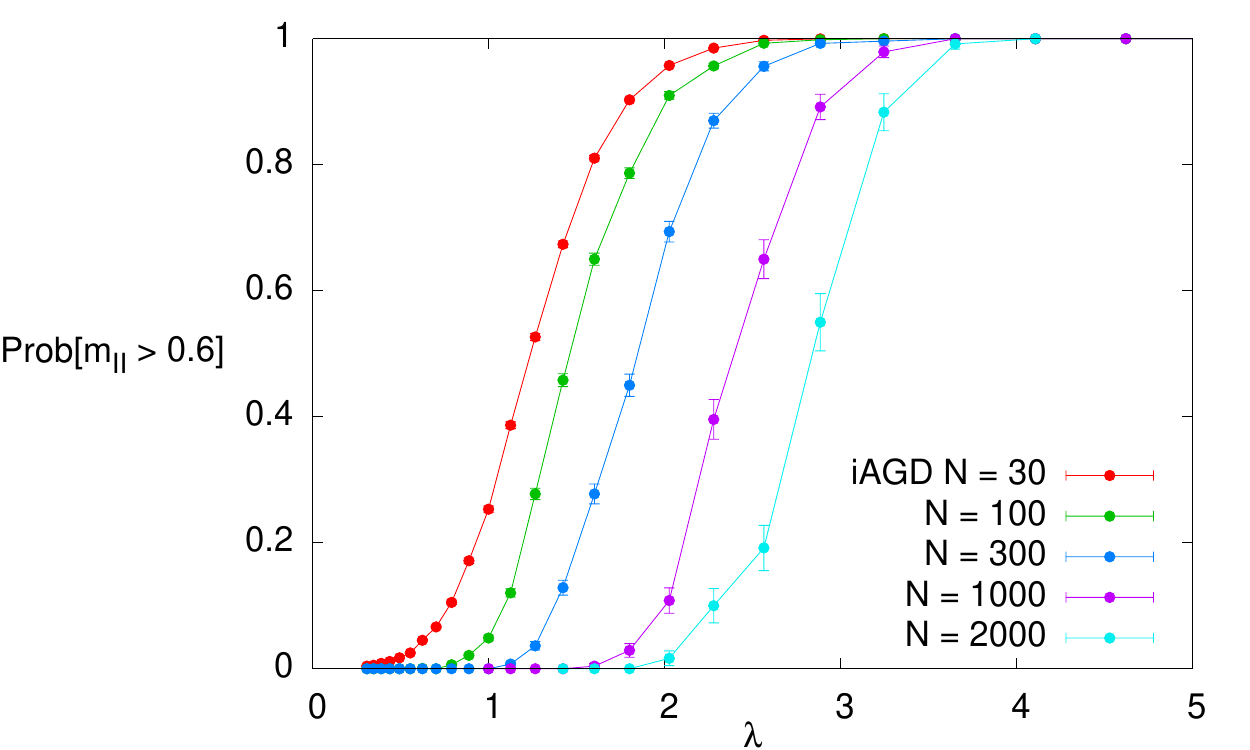}
  \hfill
  \includegraphics[width=0.48\textwidth]{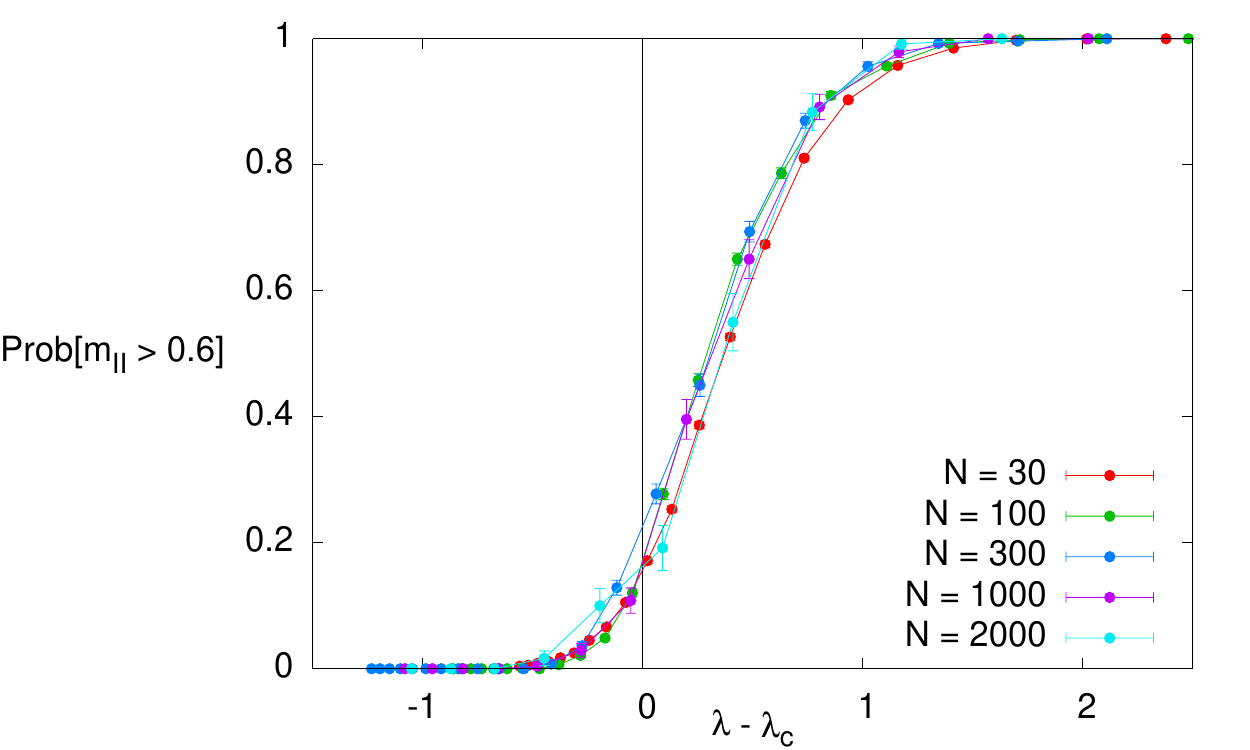}
  \caption{The most meaningful parameter to study the discontinuous phase transition leading to signal detection is the probability of being in the upper cloud of points in Figure~\ref{fig3}. We consider the iAGD algorithm and show such a probability as a function of $\lambda$ in the left panel and as a function of $\lambda - \lambda_c$ with $\lambda_c=0.37 N^{1/4}$ in the right panel. \label{fig4}}
\end{figure}

We have performed a better analysis of the data shown in Figure~\ref{fig3} by computing the probability of being in the upper cloud of points, the one corresponding to signal detection in the large $N$ limit. In practice we set a threshold value at 0.6 and compute $\text{Prob}[m_{II}>0.6]$. We show in Figure~\ref{fig4} the results of such analysis for the iAGD algorithm, together with the proper statistical errors. In the left panel we plot the probability of detecting the signal as a function of $\lambda$: since the IT threshold is $\lambda_{IT}\simeq 2.95545$ in the large $N$ limit we notice that our algorithm is still performing very efficiently on these sizes. In the right panel we show the same probabilities as a function of the critical scaling variable $\lambda - \lambda_c$ with $\lambda_c = 0.37 N^{1/4}$ and we observe a perfect data collapse within errorbars (only data for $N=30$ show tiny finite size effects).

\section*{SM7: Numerical results for finite $R$}
In hard problems other than Tensor PCA it might be impossible to work with expectation values rather than empirical averages over a finite number $R$ of replicas. Beforehand in the SM it has been discussed what are the performances expected from SAGD, here we show instead numerical results of the implementation of AGD and we compare them with those of iAGD. 
\begin{figure}
  \centering
  \includegraphics[width=0.49\textwidth]{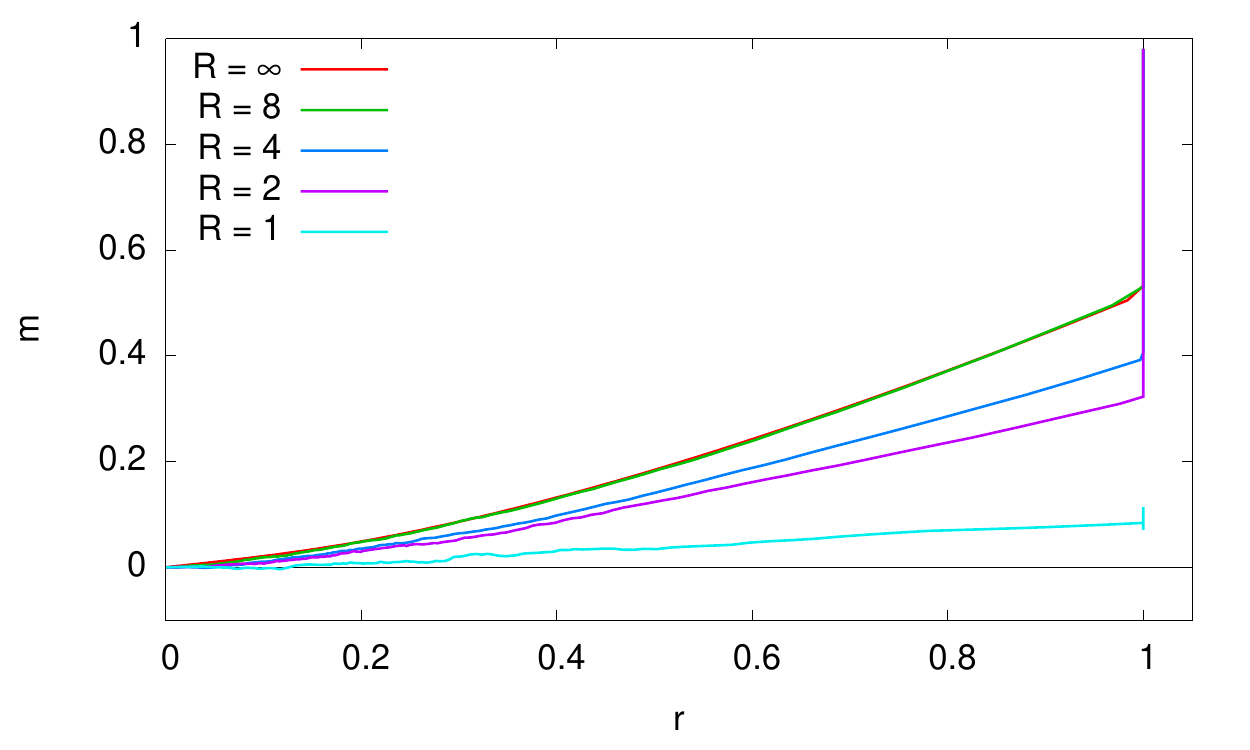}
  \includegraphics[width=0.49\textwidth]{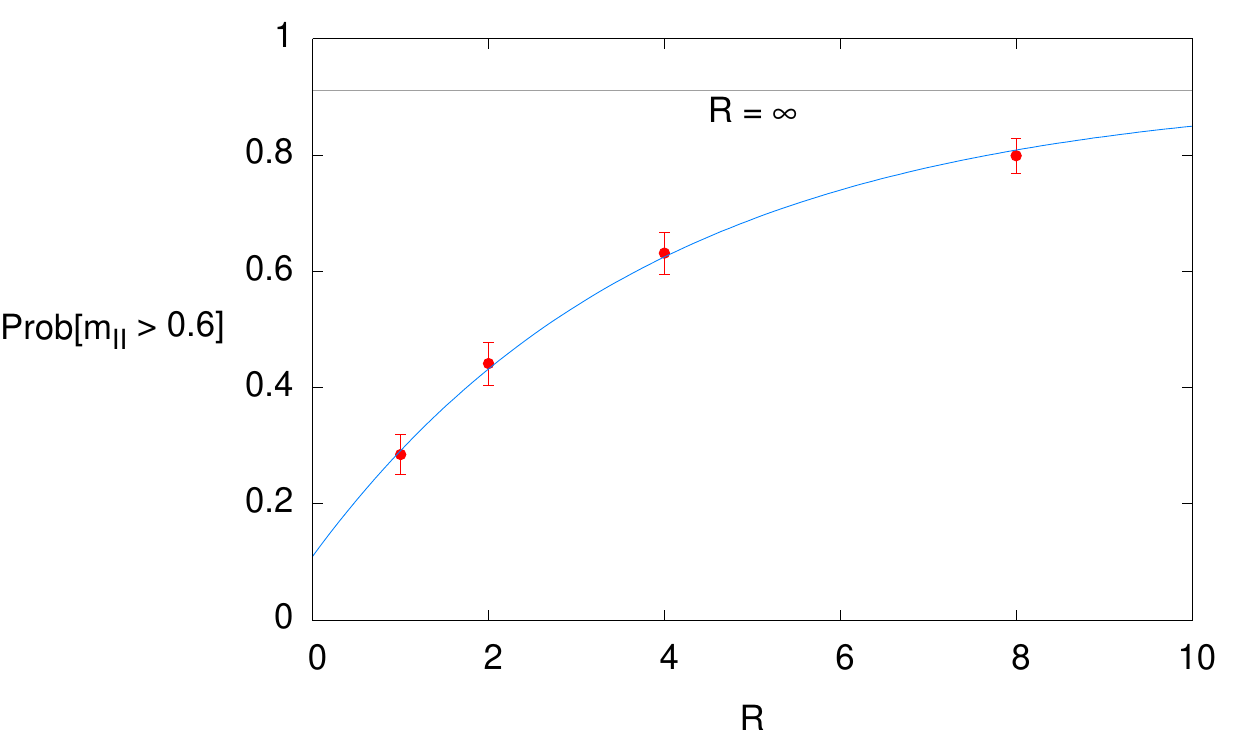}
  \caption{AGD results for $N=10^3$ and $\lambda=3$ obtained with several values of $R$ from $R=1$, {\it i.e.} GD, to $R=8$. Results from iAGD ($R=\infty$) are also added for comparison. Left: results for a given representative sample show how the AGD algorithm with moderate $R$ values already works well and may be very close to iAGD. Right: the probability of success, averaged over 180 samples, converges to the $R=\infty$ limit (shown by a horizontal line) exponentially fast (blue curve).  \label{fig5}}
\end{figure}
In Figure \ref{fig5} (left) we show typical trajectories followed by the center of mass during the execution of the AGD algorithm solving problems of size $N=10^3$ with $\lambda=3$: we plot the overlap of the center of mass with the signal $m=(\bxCM,\bv)/N$ versus the normalized norm of the center of mass $\rCM=||\bxCM||_2/\sqrt{N}$.
Recall that when $\rCM=1$, AGD and iAGD reduce to standard GD.
Observing the plot it becomes evident that a very limited number of replicas in AGD is enough to approach the behavior of iAGD (e.g., for this specific sample, AGD with $R=8$ practically matches performances of iAGD). 
In order to make a more quantitative statement we have run AGD and iAGD on 180 samples of size $N=10^3$ and $\lambda=3$. We report the results in Figure \ref{fig5} (right). The success probability grows with $R$ approaching the asymptotic value (obtained with iAGD and shown with a horizontal line) exponentially fast in $R$: the blue interpolating curve is $0.91-0.8 \exp(-R/3.9)$.

\section*{SM8: Comparison between Averaged Gradient Descent and Homotopy}

Finally we find interesting to compare the results of SiAGD with the Homotopy based algorithm, known to be the best available algorithm for Tensor PCA \cite{AnGeHK14}.
\begin{figure}
  \centering
  \includegraphics[width=0.7\textwidth]{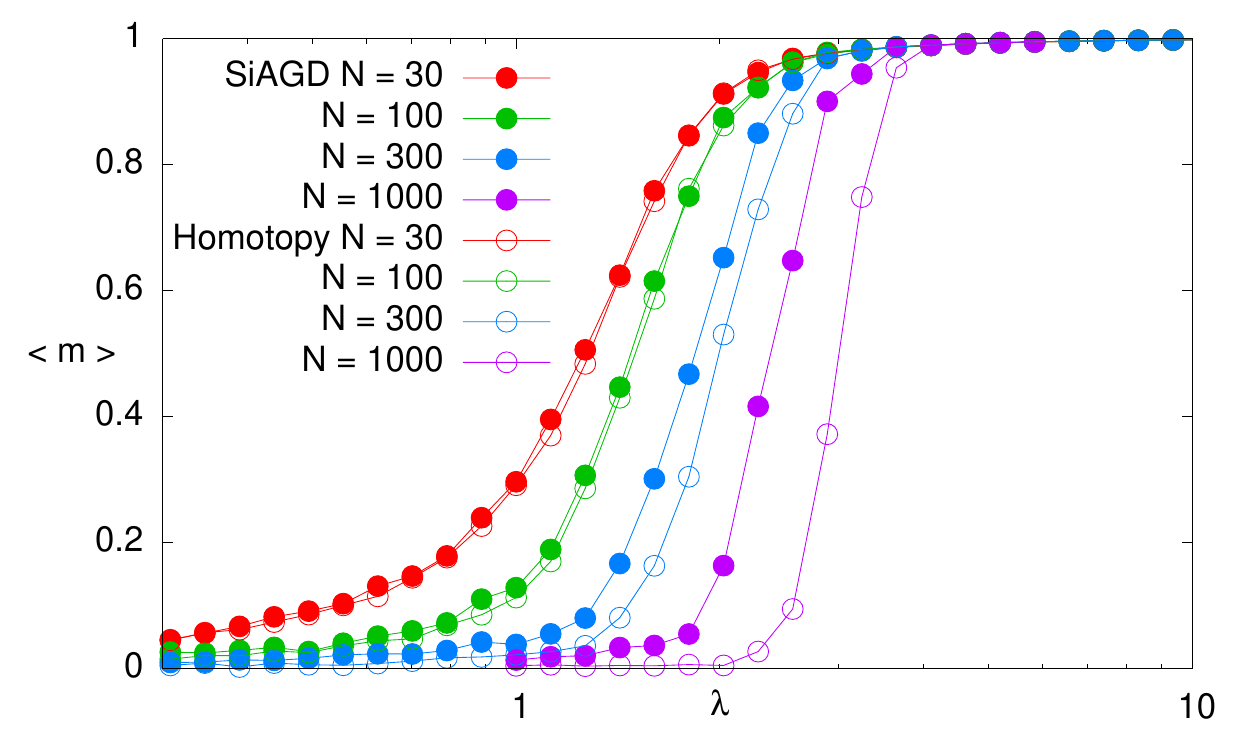}
  \caption{Comparison between Averaged Gradient Descent (with learning rate $\eta=0.125$) and Homotopy based algorithm \cite{AnGeHK14}. SiAGD achieves significantly better performances for problems with larger $N$. \label{fig6}}
\end{figure}
Remember that the second part of the SiAGD algorithm turns out to be very similar to the Homotopy algorithms as they share the same initial condition and both proceed with GD based moves.
However, while the former works with very small $\eta$ values\footnote{We used $\eta=0.125$ in most of our runs after having checked that it provides the same results a smaller $\eta$ value would return.}, the latter is implemented as if it were a GD with $\eta=\infty$.
This choice is far from optimal because such a large value for $\eta$ implies problems of convergence.
Indeed Figure~\ref{fig6} shows that the two algorithms are comparable for small $N$ but SiAGD achieves significantly better performances for problems with larger $N$.


\end{document}